\documentclass{ecai} 

\usepackage{latexsym}
\usepackage{amssymb}
\usepackage{amsmath}
\usepackage{amsthm}
\usepackage{booktabs}
\usepackage{graphicx}
\usepackage{color}
\usepackage[inline]{enumitem}
\usepackage{subcaption}

\newtheorem{definition}{Definition}

\newcommand{\BibTeX}{B\kern-.05em{\sc i\kern-.025em b}\kern-.08em\TeX}

\begin{document}

\begin{frontmatter}

\paperid{367} 

\title{What model does MuZero learn?}

\author[A]{\fnms{Jinke}~\snm{He}\thanks{Corresponding Author. Email: J.He-4@tudelft.nl.}}
\author[B]{\fnms{Thomas M.}~\snm{Moerland}}
\author[A]{\fnms{Joery A.}~\snm{de Vries}}
\author[A]{\fnms{Frans A.}~\snm{Oliehoek}}
\address[A]{Delft University of Technology}
\address[B]{Leiden University}

\begin{abstract}
Model-based reinforcement learning (MBRL) has drawn considerable interest in recent years, given its promise to improve sample efficiency. 
Moreover, when using deep-learned models, it is possible to learn compact and generalizable models from data. 
In this work, we study MuZero, a state-of-the-art deep model-based reinforcement learning algorithm that distinguishes itself from existing algorithms by learning a value-equivalent model. 
Despite MuZero's success and impact in the field of MBRL, existing literature has not thoroughly addressed why MuZero performs so well in practice.
Specifically, there is a lack of in-depth investigation into the value-equivalent model learned by MuZero and 
its effectiveness in model-based credit assignment and policy improvement, which is vital for achieving sample efficiency in MBRL. 
To fill this gap, we explore two fundamental questions through our empirical analysis: 1) to what extent does MuZero achieve its learning objective of a value-equivalent model, and 2) how useful are these models for policy improvement? 
Our findings reveal that MuZero's model struggles to generalize when evaluating unseen policies, which limits its capacity for additional policy improvement. 
However, MuZero’s incorporation of the policy prior in MCTS alleviates this problem, which biases the search towards actions where the model is more accurate. 

\end{abstract}

\end{frontmatter}

\section{Introduction}

In recent years, deep reinforcement learning (DRL) \citep{francois-lavet_introduction_2018,sutton_dyna_1991} has achieved remarkable progress, finding applications in a variety of real-world problems such as video compression \citep{mandhane_muzero_2022}, chip design \citep{mirhoseini_graph_2021}, inventory management \citep{madeka_deep_2022} and plasma control for nuclear fusion \citep{degrave_magnetic_2022}.
However, despite these advancements, sample inefficiency remains a significant obstacle that limits the broader applicability of deep reinforcement learning in practical settings. 

Model-based reinforcement learning (MBRL) \citep{moerland_model-based_2023} addresses sample inefficiency in DRL by learning predictive models of the environment. 
In the typical RL cycle, an agent interacts with the environment ({acting}) and uses the collected data to refine its policy ({learning}).
Accordingly, MBRL methods fall into two non-mutually exclusive categories \citep{van_hasselt_when_2019}: 
\begin{enumerate*}[label=(\arabic*)]
\item those that use the learned model to improve acting and
\item those that use the learned model to improve learning 
\end{enumerate*}.

One notable example of the first category is model-based exploration.
In environments where rewards are sparse, shallow exploration techniques such as epsilon-greedy exploration often fail. 
Model-based exploration addresses this by not only allowing the agent to use model prediction errors or model uncertainty as intrinsic learning signals, but also enabling more effective exploration of "interesting" regions of the environment by "planning to explore" \citep{pathak_curiosity-driven_2017,henaff_explicit_2019,shyam_model-based_2019,pathak_self-supervised_2019,sekar_planning_2020,brafman_r-max_2002,lowrey_plan_2018}.
In addition, decision-time planning methods such as Monte Carlo Tree Search (MCTS) \citep{browne_survey_2012} compute local policies online by planning with the model. 
A representative class of methods that use MCTS for decision-time planning is AlphaZero \citep{silver_general_2018,silver_mastering_2016,silver_mastering_2017}, which defeated a human Go world champion for the first time in human history. 

The other class of MBRL methods aims to improve the agent's policy and value functions without consuming additional data through model-based credit assignment.
At a high level, these methods generate synthetic data through the learned model to simulate potential outcomes of specific actions or policies.
This synthetic data is then used to update the agent's value estimates and improve the policy with a policy improvement operator.
Notably, the DYNA architecture \citep{sutton_dyna_1991} treats synthetic data as real data, integrating it into model-free learning algorithms, whereas the Dreamer methods \citep{hafner_mastering_2023,hafner_dream_2019,hafner_mastering_2020} use synthetic data exclusively for policy computation. 
In addition to using synthetic data directly, AlphaZero employs MCTS to compute a refined local policy and improved value estimates, which then serve as better learning targets for policy and value functions in replacement of model-free targets. 
Furthermore, \citet{van_hasselt_when_2019} advocate for backward planning, which assigns credits to hypothetical states through a learned inverse model. 
They argue that planning backward for credit assignment can be more robust to model errors than planning forward, as updating fictional states can be less harmful than updating real states with fictional values. 

Model-based exploration and credit assignment improve the data efficiency of RL methods by collecting more useful data and extracting more information from it. 
Beyond these two categories, learning a predictive model of the environment can also serve as an auxiliary task for representation learning \citep{hessel_muesli_2022,jaderberg_reinforcement_2022}.
In addition, in model-free planning, differentiable computation graphs that resemble the structure of planning with a model (i.e., implicit models) have been found useful as architectural priors for value and policy functions, demonstrating improved performance in combinatorial planning domains while trained via model-free losses \citep{oh_value_2017,tamar_value_2016,farquhar_treeqn_2018,guez_learning_2018,guez_investigation_2019}.

Unlike tabular methods that learn the dynamics and reward for each state-action pair, deep model-based RL (DMBRL) approaches typically learn a state representation, on top of which dynamics and reward functions are estimated. 
However, determining what relevant information to include in these state representations and how to effectively learn them remains challenging. Namely, not all features of the observations are relevant \citep{li_towards_2006}, and missing relevant features may lead to history dependence \citep{mccallum_reinforcement_1996}. 
A prevalent approach within DMBRL for learning these state representations is to model the next observation \cite{kaiser_model_2020,ha_recurrent_2018,hafner_dream_2019,hafner_mastering_2020,hafner_mastering_2023}.
However, accurately predicting high-dimensional observations, such as images, requires considerable effort in designing and training high-capacity neural networks with effective inductive biases \citep{kaiser_model_2020}.
This challenge has long been a barrier for MBRL methods, as the efficacy of model-based policy optimization strongly depends on the quality of these representations. 
Moreover, this approach often wastes significant representational power and training resources on encoding task-irrelevant information within the state representations, causing inefficiency in learning.

One approach to addressing this challenge is the development of value-equivalent models \citep{silver_predictron_2017,grimm_value_2020,grimm_proper_2021,grimm_approximate_2022,farquhar_treeqn_2018,tamar_value_2016,oh_value_2017}.
These models are specifically trained to predict the (multi-step) Bellman update, focusing solely on value-relevant aspects of the task dynamics without needing to reconstruct any observation. 
MuZero \citep{schrittwieser_mastering_2020}, a well-known MBRL algorithm, exemplifies this approach by achieving state-of-the-art performance in Atari games \citep{bellemare_arcade_2013} and superhuman performance in the game of Go, Chess, and Shogi. 

MuZero inherits much of its structure from its predecessor, AlphaZero, which uses MCTS guided by both a learned policy network and a learned value network to make decisions and generate learning targets.
However, MuZero distinguishes itself by integrating a model learned jointly with the value and policy networks, contrasting AlphaZero's use of a ground-truth model for simulation and search. 
Importantly, MuZero's model is not trained to predict the next state or observation, but instead focuses on predicting task-relevant quantities such as future rewards, policies, and values. 
This approach of learning implicit models sets MuZero apart from traditional MBRL algorithms and promises a shift towards more efficient model learning. 
Despite MuZero's empirical success and its considerable impact on MBRL \citep{antonoglou_planning_2021,hubert_learning_2021,mandhane_muzero_2022,ye_mastering_2021}, a recent study by \citep{de_vries_visualizing_2021} shows that MuZero's dynamics model can diverge significantly from real transitions, highlighting a gap in our understanding of how these models function and their efficacy in model-based credit assignment and planning.
This discrepancy underscores the necessity for a detailed investigation into the capabilities and limitations of value-equivalent models within MuZero, which motivates this work. 

The most relevant study in this direction is by \citep{hamrick_role_2022}, who studied the role of planning in MuZero's learning. 
They found that planning primarily boosts the policy network's learning by generating more informative data and constructing better training targets. Surprisingly, their findings also reveal that, in most domains, planning at evaluation time does not significantly improve performance compared to using the policy network alone, even with large search budgets. One explanation is that the policy network has converged to the optimal policy, rendering planning less useful.
Furthermore, \citet{danihelka_policy_2021} show that improving MuZero's planning enables strong performance even with extremely small search budgets ($n=2$, or $n=4$) in Go. 
This raises the question, what is the contribution of MuZero's learned model.

In this study, we aim to bridge the gap in our understanding of MuZero's learned model by exploring two fundamental questions: 
\begin{enumerate}
    \item To what extent does MuZero learn a value-equivalent model? 
    \item To what extent does MuZero learn a model that supports effective policy improvement (through planning)?
\end{enumerate}
Learning a truly value-equivalent model is essential for model-based credit assignment, which directly influences the potential for improving existing policies through model-based planning.
The more effectively we can improve existing policies through planning with the learned model, the greater the sample efficiency achieved by an MBRL method.
As such, addressing these questions can help us better understand MuZero's empirical success and inform the design of future algorithms or extensions. 

Through our empirical analysis, we find that MuZero's learned model is generally not accurate enough for policy evaluation, and the accuracy of the model decreases as the policy to evaluate deviates further from MuZero's data collection policy.
Consequently, this limits the extent to which we can find a good policy via planning.
However, we find that MuZero's incorporation of the policy prior in MCTS alleviates this problem, which biases the search towards actions where the model is more accurate. 
Based on these findings, we speculate that the role of the model in MuZero may be similar to that in model-free planning, providing a more powerful representation of value and policy functions, as the extent to which it can support policy improvement is rather limited.
Moreover, using the policy prior indirectly takes model uncertainty into account during planning, which results in a form of regularized policy optimization with the learned model.  

In Section \ref{sec_background}, we introduce the essential background. Section \ref{sec_policy_evaluation_experiments} studies trained MuZero models in the policy {\it evaluation} setting (the objective for which the models were trained).  Section \ref{sec_policy_improvement_experiments} extends the analysis to the policy {\it improvement} setting (planning). Finally, Section \ref{sec_discussion} discusses the limitations and outlook of this work.

\section{Background}\label{sec_background}
\subsection{Markov Decision Processes}
A Markov decision process (MDP) is a model that 
describes the
interaction between a decision-making agent and an environment. 
Formally, an infinite-horizon discounted MDP is a \(6\)-tuple \(\mathcal{M} = (\mathcal{S}, \mathcal{A}, \mathcal{T}, \mathcal{R}, \mu, \gamma)\) where \(\mathcal{S}\) represents the state space of the environment and \(\mathcal{A}\) denotes the set of actions that the agent can take. The initial state of the environment, \(s_0\), follows a distribution \(\mu \in \Delta(\mathcal{S})\). 
At each time step \(t\), the agent observes the environment state \(s_t\) and selects an action \(a_t\), causing the environment to transition to a new state \(s_{t+1} \sim \mathcal{T}(\cdot | s_t, a_t)\) and return a numerical reward \(r_{t} = \mathcal{R}(s_t, a_t)\). 
Given a stationary policy \(\pi: \mathcal{S} \rightarrow \Delta(\mathcal{A})\), the value function \(V\) measures the expected discounted sum of future rewards from a state following \(\pi\) afterward, \(V^{\pi}(s) = E[ \sum_{t=0}^{\infty} \gamma^{t} r_t | \pi, s_0 = s]\), where \(\gamma \in [0,1)\) is the discount factor. 
In this work, we specifically focus on environments with deterministic dynamics, where the next state \(s_{t+1}\) is solely determined by the current state \(s_t\) and action \(a_t\) with probability \(1\): \(s_{t+1} = \mathcal{T}(s_t, a_t)\). 

\subsection{MuZero}
\label{sec:background-muzero}
MuZero \citep{schrittwieser_mastering_2020} is a recent MBRL method that has achieved state-of-the-art performance in Atari games and matched the superhuman performance of AlphaZero \citep{silver_general_2018} in Go, chess, and shogi.

\textbf{Components} Similar to other MBRL methods, MuZero learns a deterministic world model that consists of a representation function $h_\theta$ and a dynamics function \(g_\theta\). 
The representation function encodes an environment state \(s_t\) into a latent state \(z_t^0 = h_\theta(s_t)\). 
Here, the subscript denotes the real time step at which the encoding occurs, and the superscript denotes the number of time steps that have been spent in the learned model since then. 
Given a latent state \(z_t^{k}\) and an action \(a_{t+k}\), the dynamics function predicts the next latent state \(z_t^{k+1}\) and the reward \(u_{t}^{k}\), \((z_t^{k+1}, u_{t}^{k}) = g_\theta(z_t^k, a_{t+k})\). 
Apart from the representation and dynamics functions, MuZero uses a prediction function to predict the value and policy at a latent state, \(\pi_t^k, v_t^k = f_\theta(z_t^k)\).
For ease of notation, we split the prediction function into a policy function \(\pi_\theta(z_t^k)\) and a value function \(v_\theta(z_t^k)\). 
To distinguish between the policy function \(\pi_\theta\) and the MuZero policy \(\pi^{\text{MuZero}}\), which runs MCTS, we will refer to the former as the policy prior and the latter as MuZero's behavior policy. 

\textbf{Acting} MuZero makes decisions by planning with the learned model. At each time step \(t\), MuZero encodes the environment state \(s_t\) into a latent state \(z_t^0\) and uses it as the root node to perform MCTS. As the result of the search, MuZero selects the action \(a_t\) by sampling from a distribution that is constructed using the visit counts at the root node and a temperature parameter  \(T \in (0, \infty) \):
\begin{equation}\label{equation:muzero_acting}
    \pi^{\text{MuZero}}(a | s_t) = \frac{N(z_t^0, a)^{1/T}}{\sum_b{N(z_t^0, b)^{1/T}}}
\end{equation}
MuZero's planning differs from traditional MCTS methods like UCT \citep{kocsis_bandit_2006} in two key ways. 
First, rather than employing random rollouts for leaf node value estimation, MuZero uses its learned value function \(v_\theta\). 
Second, MuZero integrates the policy prior into its action selection, guiding the simulation of actions at tree nodes:
\begin{equation}
    \arg \max_a \big[Q(z,a) + c \cdot \pi_\theta(a|z) \cdot \frac{\sqrt{\sum_b N(z,b)}}{1+N(z,a)} \big]
\end{equation}
where \(c = c_1 + \log{(\frac{\sum_b N(z,b) + c_2 + 1}{c_2})}\) with \(c_1 = 1.25\) and \(c_2 = 19652\). \(Q\) and \(N\) are the estimated values and visit counts of actions. MuZero inherits much of this search mechanism from AlphaZero, including the use of policy prior and value functions. A recent study by \citet{grill_monte-carlo_2020} shows that the use of policy prior in MCTS effectively makes the action visit distribution at the root node track the solution of a local regularized policy optimization problem. 

\begin{figure}[h]
    \centering 
    \includegraphics[width=\linewidth]{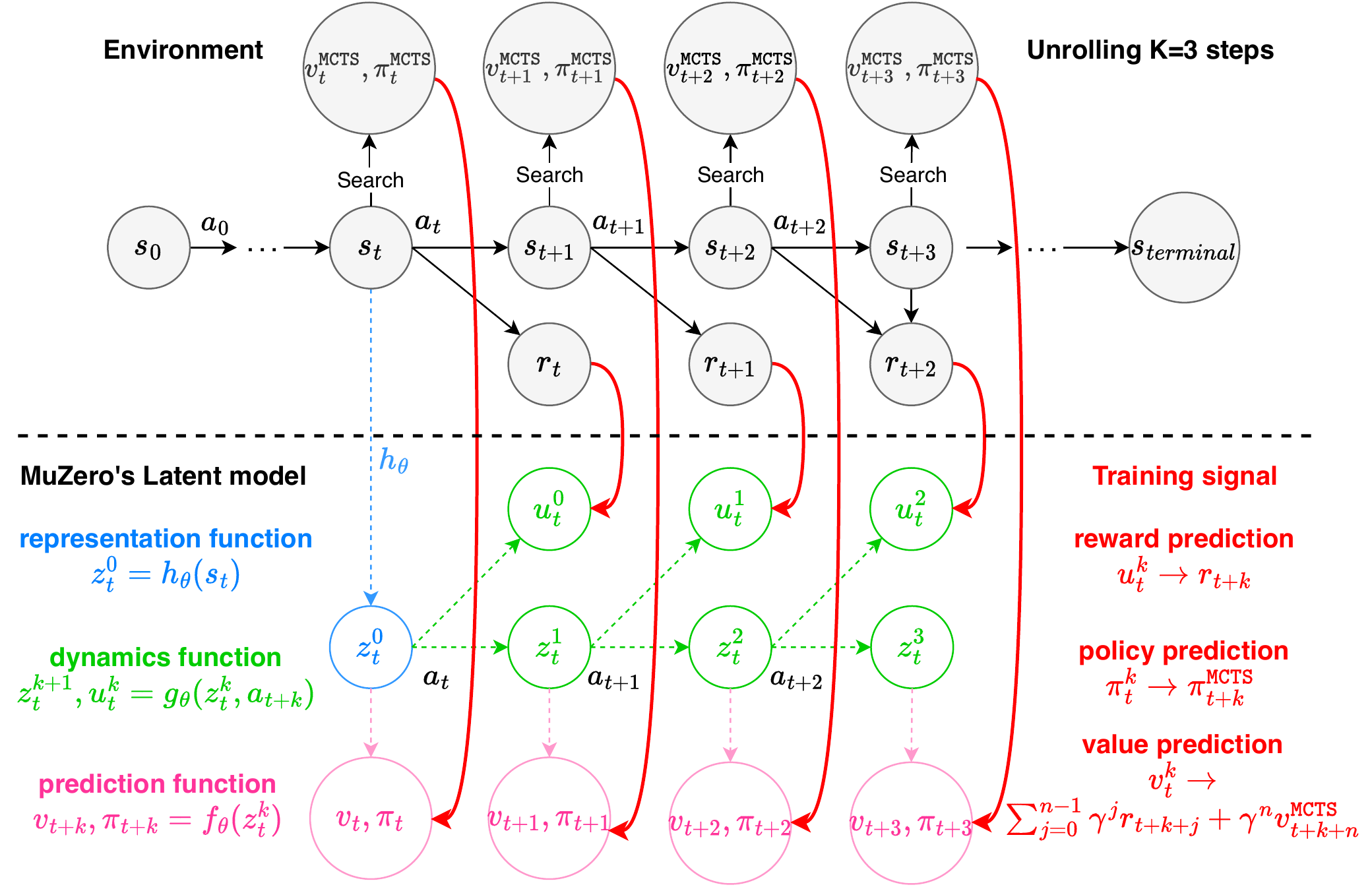}
    \vspace{-5pt}
    \caption{An illustration of MuZero's loss function.}
    \vspace{15pt}
    \label{fig:muzero_loss}
\end{figure}

\textbf{Training}
\label{sec:muzero_training}
The key difference between MuZero and many prior works in MBRL lies in their approach to learning the model.
In Figure \ref{fig:muzero_loss}, we illustrate the loss function of MuZero. 
Essentially, given a segment of a real episode that starts from state \(s_t\), MuZero unrolls its model for \(K\) simulated time steps (below the dotted line) and compares these to real experiences (above the dotted line). 
This comparison results in a loss consisting of three terms for each of the \(K\) steps. 
The first term is the per-step reward prediction loss, with the target being the real reward received. 
The second term is the policy prediction loss, with the target being the action visit distribution of the MCTS at the root node. 
The third term is the value prediction loss, with the target being the discounted sum of \(n\)-step real rewards plus a value estimation bootstrapped from MCTS \(n\) steps into the future \(v^{\text{target}}_{t} = \sum_{k=0}^{n-1}\gamma^k r_t + \gamma^n v^{\mathtt{MCTS}}_t\). 
For simplicity, we omit the links for value targets in Figure \ref{fig:muzero_loss}.
All components of MuZero are trained jointly end-to-end by minimizing the aggregated loss. 
In practice, MuZero incorporates various additional techniques from the literature, such as prioritized experience replay, to improve training. 
It also introduces a novel algorithm called `reanalyse' to generate fresh targets from old trajectories by re-running MCTS on them using the latest network. 
In our experiments, we followed the original paper and used all the aforementioned techniques to train our MuZero agents. 

\subsection{The value equivalence principle}

The value equivalence principle, introduced by \citet{grimm_value_2020,grimm_proper_2021,grimm_approximate_2022}, is motivated by the consideration that the construction of a model should take into account the final use of the model. 
It defines the order-\(k\) value equivalence class \(\mathcal{M}^k(\Pi, \mathcal{V})\) as the subset of all models that can predict the correct \(k\)-step Bellman update for any \(\pi \in \Pi\) and \(v \in \mathcal{V}\) in a set of policies \(\Pi\) and functions \(\mathcal{V}\). 
For \(k \rightarrow \infty\), an additional proper value equivalence class \(\mathcal{M}^\infty(\Pi)\) is defined, which excludes the set of functions from the specification. 
In essence, a model in the proper value equivalence class must have the true value function as the fixed point of the Bellman operators for all policies in this set. 
As such, these models can be seen as the \(Q^\mathbf{\Pi}\)-irrelevance abstractions for the MDP \citep{li_towards_2006}.
Notably, the largest proper value equivalence class that guarantees optimal planning is \(\mathcal{M}^{\infty}(\mathbf{\Pi}^{\textnormal{DET}})\), where \(\mathbf{\Pi}^{\textnormal{DET}}\) is the set of all deterministic policies. Moreover, \citet{grimm_proper_2021} show that a simplified version of MuZero's loss upper bounds the proper value equivalence loss, making an explicit connection between the theory of value equivalence principle and the strong empirical performance of MuZero. 

\section{Policy Evaluation Experiments} 
\label{sec_policy_evaluation_experiments}
We are interested in the extent to which MuZero's learned model supports additional policy improvement, which is crucial for MuZero's sample efficiency as an MBRL method. 
Our hypothesis is that, since MuZero's model is trained on data collected by previous policies, it is not generally value equivalent for all policies, especially those that have not been executed. 
As accurate policy evaluation is the basis for effective policy improvement, this will limit the extent to which MuZero can additionally improve its policy through planning.
In this section, we validate this hypothesis. 

\subsection{Training MuZero agents}
For our empirical analysis, we used three fully observable deterministic environments, as MuZero was designed for this setting: Cart Pole, a deterministic version of Lunar Lander, and Atari Breakout \citep{bellemare_arcade_2013}.
We trained \(30\) MuZero agents with different random seeds for Cart Pole and Lunar Lander, and \(20\) for  Atari Breakout. 
For each agent, we saved the model weights at different training steps.
In the figures below, we aggregate results from different seeds/agents and report their means as well as the corresponding standard errors.

In Cart Pole and Lunar Lander, we extensively trained the MuZero agents, for \(100\)K and \(1\)M steps. 
This way, we can conduct our analysis on the trained agents throughout their lifecycle of learning. 
For Atari Breakout, we adopted the same setup as EfficientZero \citep{ye_mastering_2021} but extended the training from 100K steps to 500K steps. 

\begin{figure}[h]
    \centering
    \includegraphics[width=\linewidth]{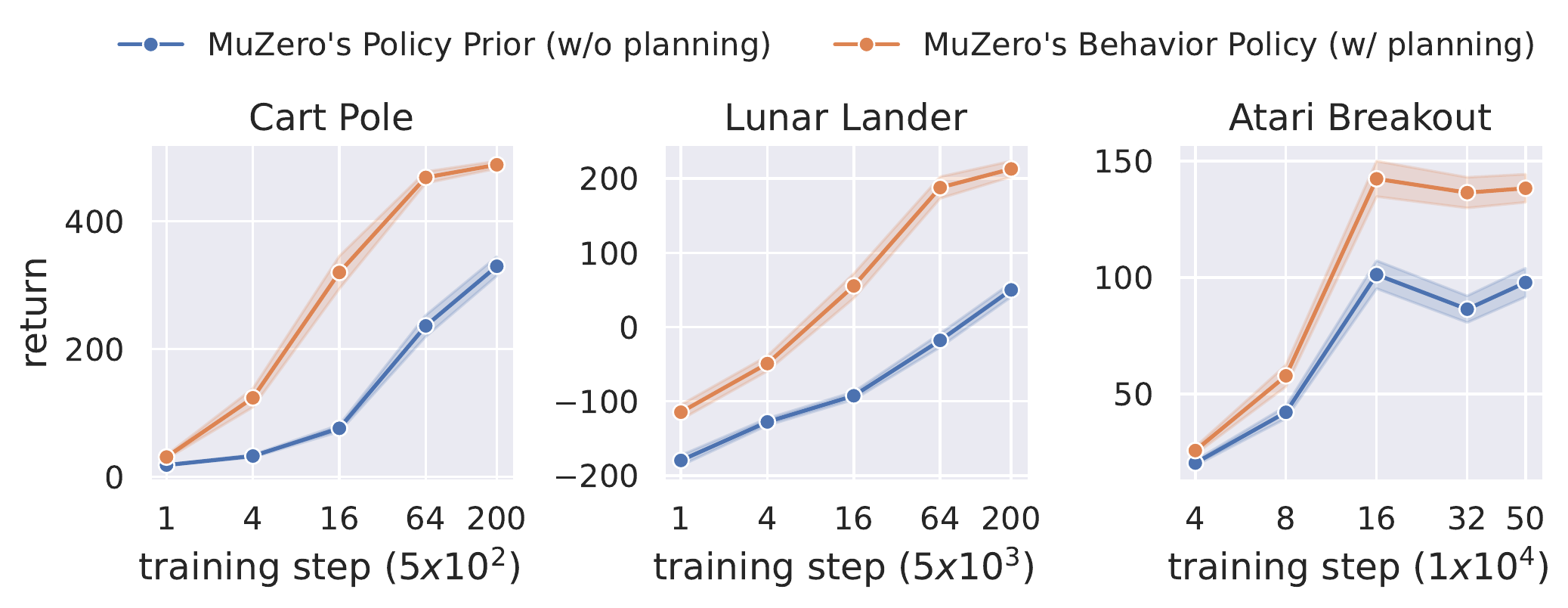}
    \vspace{-5pt}
    \caption{
    Online performance of MuZero agents in Cart Pole (Left), Lunar Lander (Middle) and Atari Breakout (Right). %
    }
    \vspace{15pt}
    \label{fig:training}
\end{figure}

Figure \ref{fig:training} shows the online performance of MuZero agents at various training steps, which make decisions by performing MCTS planning (Equation \ref{equation:muzero_acting}).
For reference, we also plot the performance of the policy prior of the same agent, which samples actions directly from the policy network, $a_t \sim \pi_\theta(\cdot|h_\theta(s_t))$.
Throughout training, planning enables MuZero to achieve a substantially better performance than directly using the policy prior. 
It is important to note that the improved performance may not only come from the learned model but may also come from the value network, which is used to estimate the value of leaf nodes in MuZero's MCTS. 
In the following,  we will take a deeper look at how much the learned model contributes to this.

\subsection{Evaluating the learned model}

To assess MuZero's learned model in the context of policy evaluation, we will compare it to the ground-truth model. Through experiments, we aim to answer questions in the format of \textit{how much value prediction error should we expect when using MuZero's learned model to evaluate a policy \(\pi\)}? 

Considering that MuZero does not train or employ the model for infinite-horizon rollouts, we impose a limit on the evaluation horizon when assessing the value prediction error. 
For each state $s_t$, we define the discounted sum of future rewards for taking an action sequence \((a_t, \ldots, a_{t+h-1})\) as:
\begin{equation}
    v^{a_{t:t+h-1}}(s_t)~=~\sum_{k=0}^{h-1} \gamma^{k} r_{t+k}
\end{equation}
where $r_{t+k} = \mathcal{R} (s_{t+k}, a_{t+k})$ and $s_{t+k+1} = \mathcal{T}(s_{t+k}, a_{t+k})$, assuming the environment is deterministic. 
As mentioned in Section \ref{sec:background-muzero}, MuZero predicts the value of this action sequence by first encoding the state into a latent state $z_t^0 = h_\theta(s_t)$ and then rolling out the model from the latent state:
\begin{equation}
    \hat{v}^{a_{t:t+h-1}}(s_t) = \sum_{k=0}^{h-1} \gamma^{k} u_{t}^k
\end{equation}
where $z_t^{k+1}, u_{t}^{k} = g_\theta(z_t^k, a_{t+k})$. Then, we can define the value prediction error of the learned model for the action sequence $a_{t:t+h-1}$.

\begin{definition}
\label{def:value_prediction_error_for_action_sequences}
    The value prediction error of using the learned model to evaluate the action sequence $a_{t:t+h-1}$ at state $s_t$ is:
    \begin{equation}
        \lvert v^{a_{t:t+h-1}}(s_t) - \hat{v}^{a_{t:t+h-1}}(s_t) \rvert
    \end{equation}
\end{definition}
As a stationary policy \(\pi: \mathcal{S} \rightarrow \Delta(\mathcal{A}) \) that operates in the original environment defines a distribution over such action sequences:
\begin{equation}
\Pr(a_{t:t+h-1} | s_t, \pi) = \prod_{k=0}^{h-1} \pi \big(a_{t+k} | s_{t+k} =\mathcal{T}(s_{t+k-1}, a_{t+k-1}) \big)
\end{equation}
We can define the value prediction error for evaluating $\pi$.
\begin{definition}
    \label{def:value_prediction_error_for_policies}
    The value prediction error of using the learned model to evaluate a stationary policy $\pi$, which operates in the original environment, for horizon $h$ is:
    \begin{align}
    \left| v^{\pi}_h(s_t) - \hat{v}^{\pi}_h(s_t) \right| &= \left| \mathbb{E}_{\substack{a_{t:t+k-1} \\ \sim \Pr(\cdot|s_t, \pi)}} \left[  v^{a_{t:t+h-1}}(s_t) - \hat{v}^{a_{t:t+h-1}}(s_t) \right] \right|
    \end{align}
\end{definition}

\subsection{How accurately can MuZero's learned model predict the value of its own behavior policy?}

As the model is trained on data collected by MuZero's behavior policy \(\pi^{\text{MuZero}}\), we expect it to be at least accurate on this data collection policy. 
Therefore, we begin our investigation by examining the model's prediction performance on this policy.
Due to the continuous state spaces of our environments, it is not feasible to enumerate all states and compute the error for each one of them.
Instead, we sample states from MuZero's on-policy state distribution \(d_{\pi^{\text{MuZero}}}\). For each sampled state, we conduct the evaluation and aggregate the errors. To facilitate tractable evaluation, at every state $s$, we use Monte Carlo sampling to estimate both the true value $v_h^{\pi^{\text{MuZero}}}(s)$ and the value predicted by MuZero's model $\hat{v}_h^{\pi^{\text{MuZero}}}(s)$. 
\begin{figure}[h]
    \centering
    \includegraphics*[width=\linewidth]{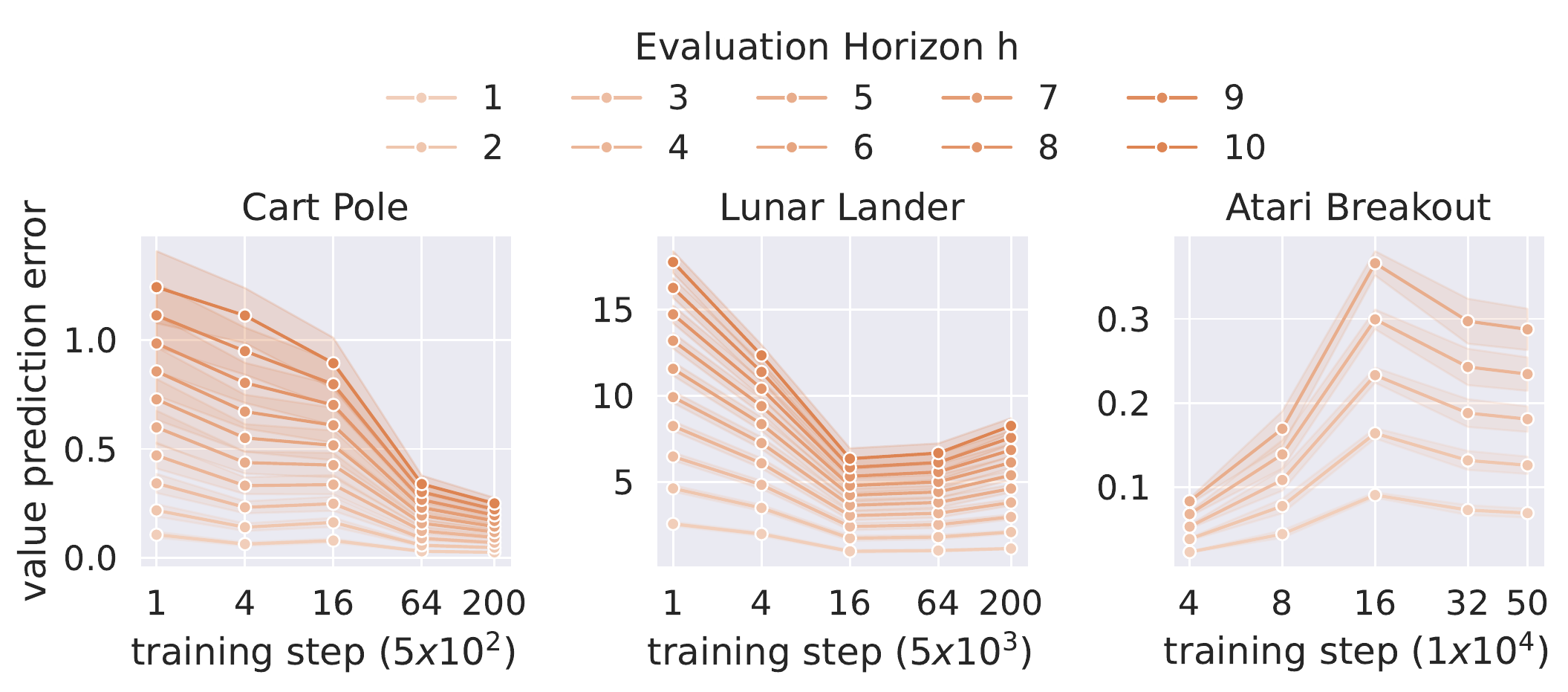}
    \vspace{-5pt}
    \caption{Value prediction error of using MuZero's learned model to evaluate its own behavior policy.}
    \vspace{15pt}
    \label{fig:value_estimation_error}
\end{figure}

\begin{figure*}[t]
    \centering
    \begin{subfigure}[b]{0.85\linewidth}
        \centering
        \includegraphics*[width=\linewidth]{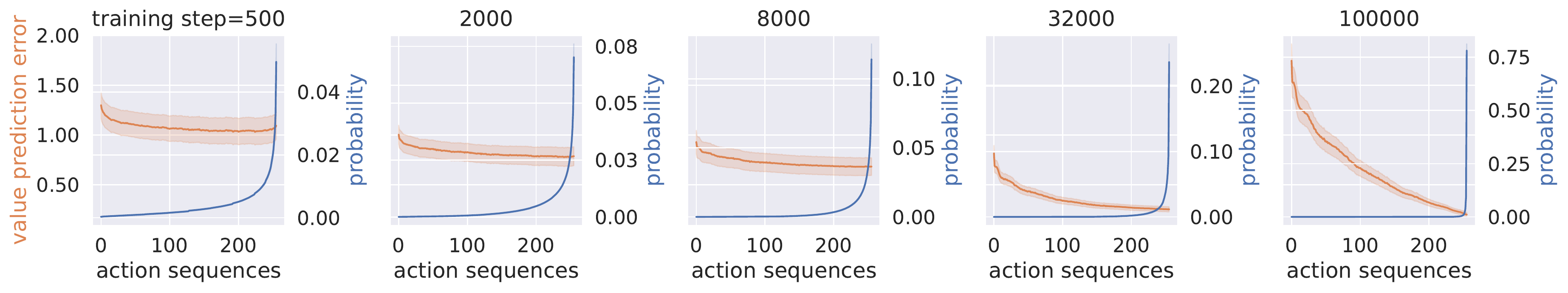}
        \vspace{-15pt}
        \caption{Cart Pole}
        \vspace{8pt}
    \end{subfigure}
    \begin{subfigure}[b]{0.85\linewidth}
        \centering
        \includegraphics*[width=\linewidth]{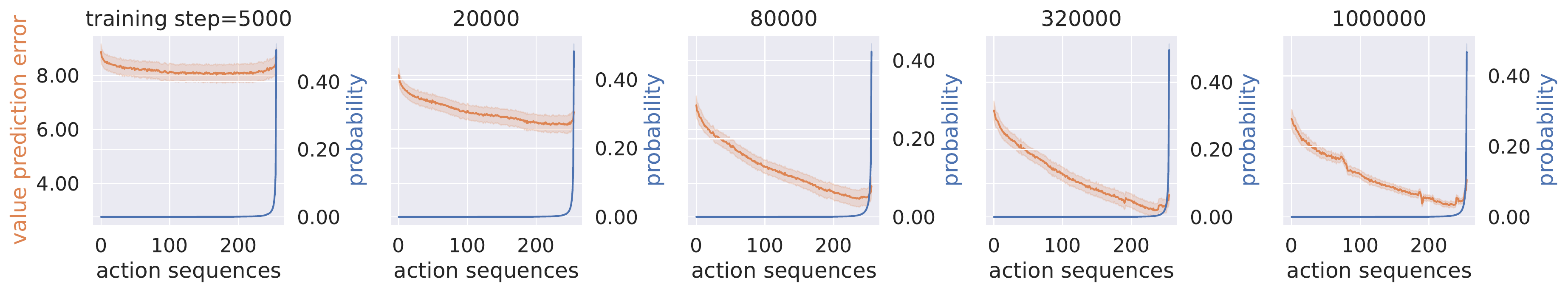}
        \vspace{-15pt}
        \caption{Lunar Lander}
        \vspace{8pt}
    \end{subfigure}
    \begin{subfigure}[b]{0.85\linewidth}
        \centering
        \includegraphics*[width=\linewidth]{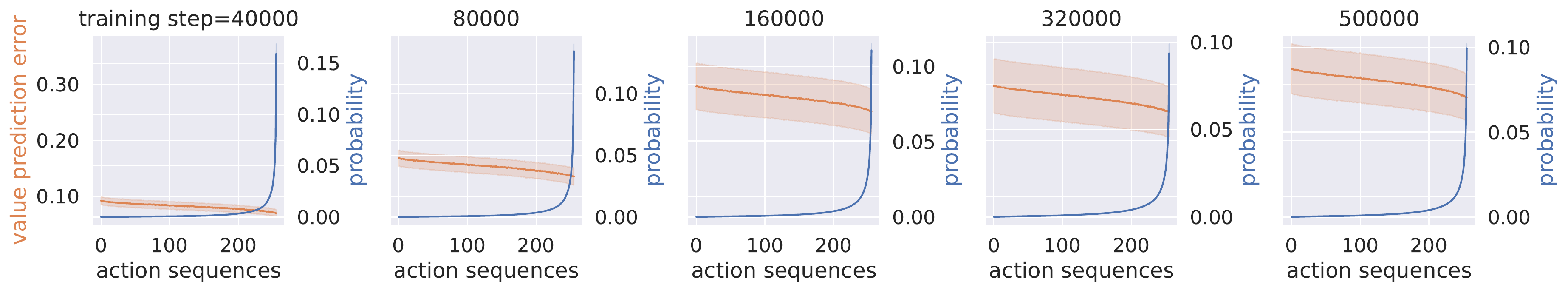}
        \vspace{-15pt}
        \caption{Atari Breakout}
        \vspace{12pt}
    \end{subfigure}
    \caption{X-axis: action sequences sorted by their probabilities of being taken by the behavior policy (from unlikely to likely reading from left to right). Y-axis: the probabilities (blue, from small to large) and the corresponding value prediction errors (yellow, from small to large).  Action sequences with higher probabilities to be taken by MuZero’s behavior policy correlate with lower value prediction errors by MuZero’s learned model. This implies that the learned model is less accurate for policies that are different from the current data collection policy.}
    \vspace{15pt}
    \label{fig:value_estimation_error_prob}
\end{figure*}

In Figure \ref{fig:value_estimation_error}, we report the value prediction error (Y-axis) under various evaluation horizons (lines) across different training steps (X-axis). The maximum evaluation horizons are set to the number of unrolling steps during training, which are \(\{10,10,5\}\) for Cart Pole, Lunar Lander, and Atari Breakout, respectively. 
In all environments, value predictions for short horizons are highly accurate, with errors approaching zero. 
However, as the evaluation horizon increases, the errors consistently grow larger.
On the one hand, this is not surprising because learned models are known prone to accumulate errors during long rollouts \citep{lambert_investigating_2022}.
On the other hand, this shows that models learned by MuZero cannot be fully value equivalent as they are not even accurate enough to predict values for the policy that collects the training data.
Errors at different training steps are generally not comparable due to the different state distributions, but the decreasing errors observed at the end of training suggest the convergence of the policy as a possible explanation.

\subsection{How accurately can MuZero's learned model evaluate policies that are different from the behavior policy?}

To assess whether MuZero's learned model effectively supports planning, we will investigate its capacity to generalize and accurately predict values beyond its own data collection policy. Our hypothesis is that the model will exhibit increasing inaccuracies when evaluating policies that differ significantly from the behavior policy, which is responsible for collecting the training data. 

To test this hypothesis, we conduct an experiment focusing on the relationship between the value prediction error for an action sequence $|v^{a_{t:t+h-1}}(s_t) - \hat{v}^{a_{t:t+h-1}}(s_t)|$ and the probability of the behavior policy selecting this action sequence $\Pr(a_{t:t+h-1}|s_t, \pi^{\text{MuZero}})$. 
We again sample states from MuZero's on-policy state distribution \(d_{\pi^{\text{MuZero}}}\) and compute the probabilities and value prediction errors for \textit{all} action sequences of length \(\{8, 4, 4\}\) for Cart Pole, Lunar Lander and Breakout (limited by computation budget). 
Considering that both probabilities and errors are real-valued and non-uniformly distributed, we aggregate results as follows: first, for each state, we rank action sequences by their probabilities of being chosen by MuZero's behavior policy. Then, we compile statistics for action sequences with the same ranks. 
Finally, we combine results from different agents and report their means and standard errors. 

In Figure \ref{fig:value_estimation_error_prob}, we present the results.
Here, the X-axis represents the action sequences that are sorted by their likelihood of being taken by the current behavior policy, ranging from unlikely to likely.
On the Y-axis, we plot both the probabilities (blue) and the corresponding value prediction errors (yellow).
The results clearly show that, as the likelihood of the behavior policy selecting the action sequence decreases, the value prediction error for that action sequence increases. 
This trend is consistently observed across different environments and training steps. 
Moreover, it seems to become more evident with more training, possibly because the behavior policy becomes more deterministic. 
This finding supports our hypothesis that the model is more reliable in predicting values for the behavior policy, which collects the training data.
Consequently, evaluating a policy using MuZero's learned model may yield increasingly inaccurate results as the policy to evaluate deviates further from the behavior policy.

\begin{figure*}[t]
    \centering
    \begin{subfigure}[t]{0.33\linewidth}
        \centering
        \includegraphics*[width=\linewidth]{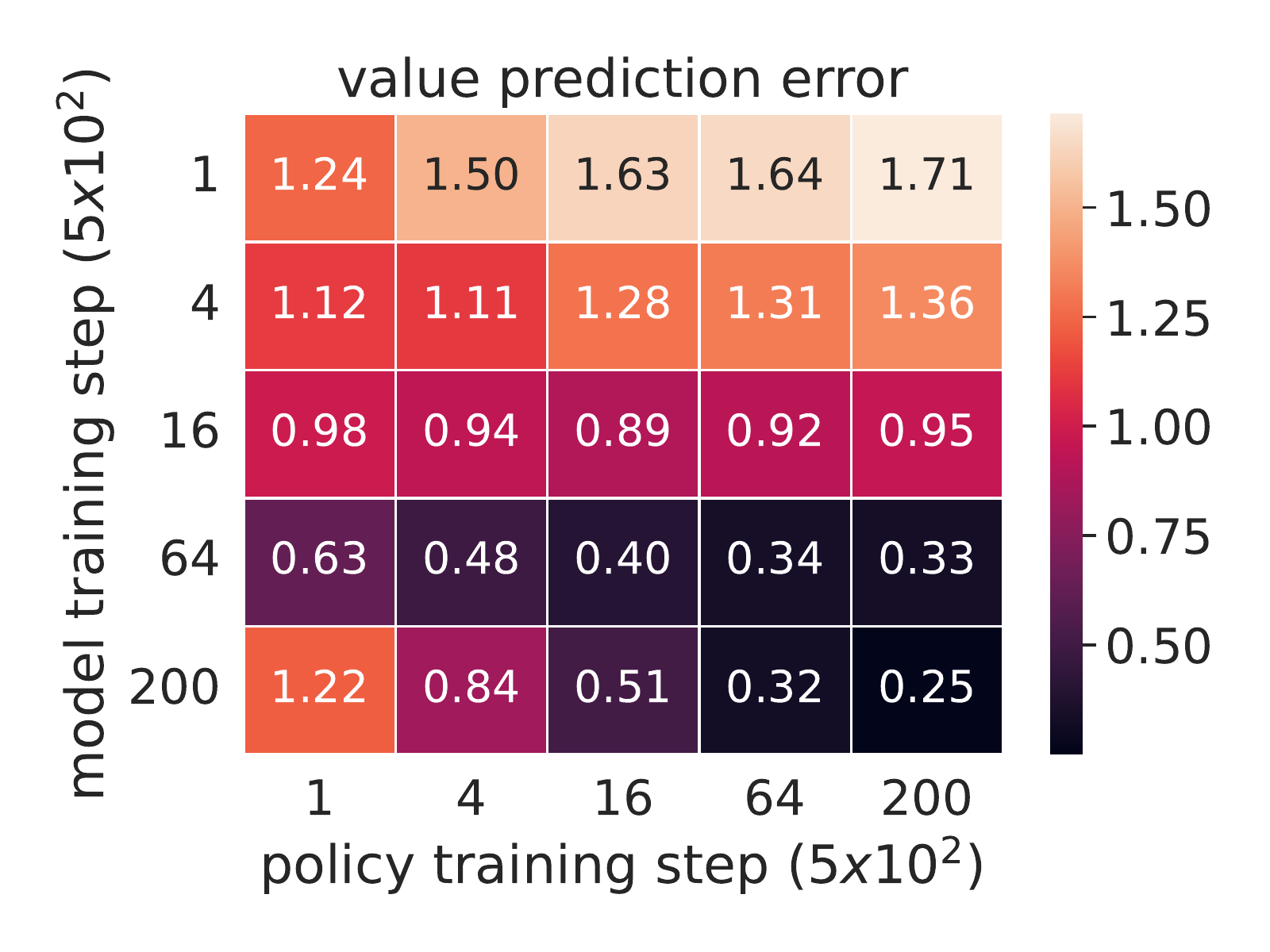}
         \vspace{-15pt}
        \caption{Cart Pole}
    \end{subfigure}
    \begin{subfigure}[t]{0.33\linewidth}
        \centering
        \includegraphics*[width=\linewidth]{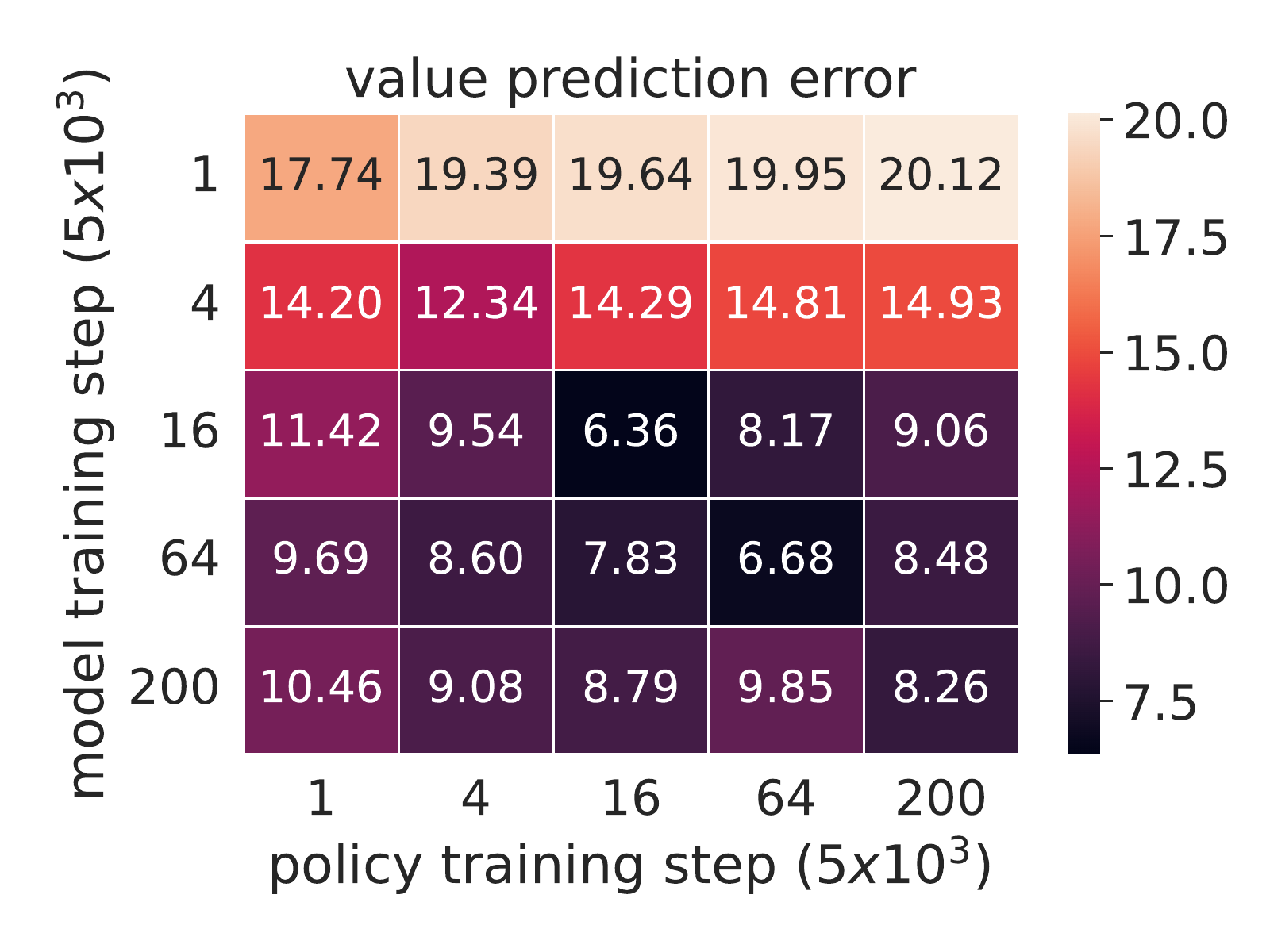}
        \vspace{-15pt}
        \caption{Lunar Lander}
    \end{subfigure}
    \begin{subfigure}[t]{0.33\linewidth}
        \centering
        \includegraphics*[width=\linewidth]{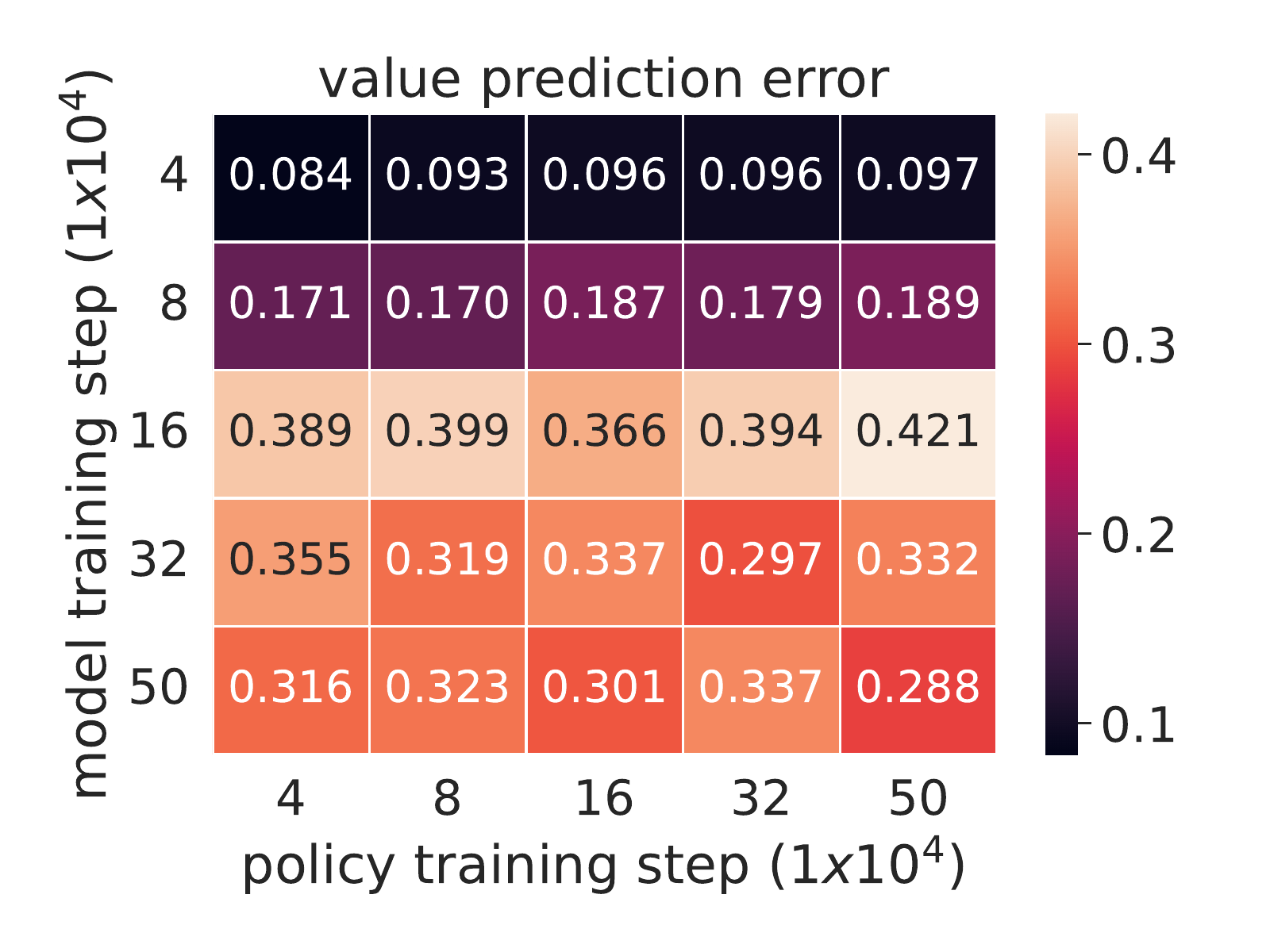}
         \vspace{-15pt}
    \caption{Atari Breakout}
    \vspace{15pt}
    \end{subfigure}
        \caption{Cross model policy evaluation. We evaluate MuZero's behavior policy at training step Y (column) with the learned model at training step X (row) and measure the value prediction error. Results are aggregated over states sampled from MuZero's on-policy state distribution at training step X (same as the model). 
    }
    \vspace{10pt}
    \label{fig:cross_model_policy_evaluation}
\end{figure*}

\subsection{How accurately can MuZero's learned model from one training step evaluate the behavior policy of the same agent from another training step?}
\label{sec:cross_model_policy_eval}
In our previous experiment, we investigated the generalization capability of the learned model in predicting values for policies that differ from the behavior policy. 
We accomplished this by considering all action sequences of length $h$ at each state, possibly including actions that are unlikely to be taken by any sensible policy.
In this experiment, we conduct a similar analysis but with a focus on more interesting policies. 
Specifically, we assess the accuracy with which the model at training step X (row) can evaluate the policy (of the same agent) at training step Y (column). 
The idea is that if a model cannot accurately evaluate high-performing future policies, planning with it would not be effective in finding a good policy. For this experiment, we set the evaluation horizon for each environment to the number of unrolling steps during training. Results are aggregated over states sampled from MuZero's on-policy state distribution at training step X (same per row as the model). Errors at different model steps (rows) are not directly comparable due to the different state distributions. 

It is clear from Figure \ref{fig:cross_model_policy_evaluation} that models at all training steps are most accurate when evaluating policies at the same training steps, as the error per row is smallest at the diagonal. 
This aligns with our earlier finding: learned models are more accurate when assessing action sequences with a higher selection likelihood by the current behavior policy. 
Notably, we can see that using early models to evaluate future policies results in large errors in Cart Pole and Lunar Lander. This may have important implications for policy improvement: \emph{if the model at the current training step cannot accurately evaluate a future policy that performs better, then the extent to which we can improve our current policy will be limited.} 
Intuitively, if the model does not know a high-performing policy is good, then a deeper search in the model would not help us find the policy. 
This phenomenon is less evident in Breakout, possibly because the evaluation horizon is too short in this environment to allow for an observable difference in rewards across the policies ($5$ vs $10$ in the other environments). 

\section{Policy Improvement Experiments} \label{sec_policy_improvement_experiments}
\label{sec:planning_experiments}

Our policy evaluation experiments indicate that MuZero's learned model can become increasingly inaccurate when evaluating policies differing from the data collection policy, particularly those unseen during training. 
In this section, we investigate the natural follow-up question, the question that we are most interested in: what is the effect of this on policy improvement through planning?

Intuitively, if the agent is given the ground-truth model and has an infinite budget for planning, it can act optimally by exhaustively searching with the model. 
In this case, sample efficiency is maximized because the agent does not need any sample from the environment to learn.
However, in real-world scenarios, the planning budget is always constrained, both during training and deployment. 
To improve planning with these constraints, AlphaZero and MuZero employ a learned policy prior to guide the action simulation in MCTS.
If the policy prior is well-trained, it can greatly enhance planning, but if not, it might diminish efficiency.

We evaluate planning using MuZero's learned model both with and without the guidance of the policy prior. In the latter case, we replace the policy prior with a uniform prior, allowing for a form of ``free search''. 
To focus on evaluating the contribution of the learned model, we modify MuZero's MCTS by replacing the value network's predictions at leaf nodes with random rollouts in the model, a standard approach to estimating the value of a leaf node in MCTS. 
To address the computational costs of long rollouts in neural models, we restrict the planning horizon to \( \{16, 128, 32\} \) in Cart Pole, Lunar Lander, and Atari Breakout. 
For comparison, we include the baselines of the policy prior and planning with the ground-truth model using the same MCTS.
The goal of this experiment is to answer two questions:
\begin{enumerate*}[label=(\alph*)]
    \item How effectively does MuZero's learned model support free search?
    \item To what extent can MuZero improve its policy by planning with the learned model?
\end{enumerate*} 

\begin{figure*}[t]
    \centering
    \begin{subfigure}{0.8\linewidth}
        \centering
        \includegraphics*[width=\textwidth]{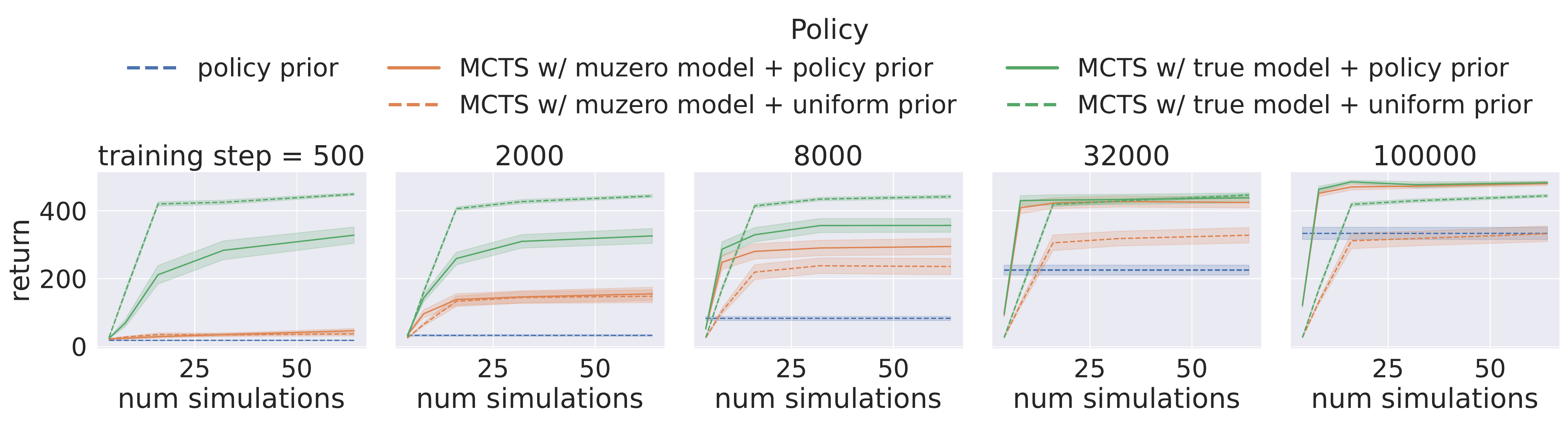}
        \vspace{-15pt}
        \caption{Cart Pole}
        \vspace{10pt}
    \end{subfigure}
    \begin{subfigure}{0.8\linewidth}
        \centering
        \includegraphics*[width=\textwidth]{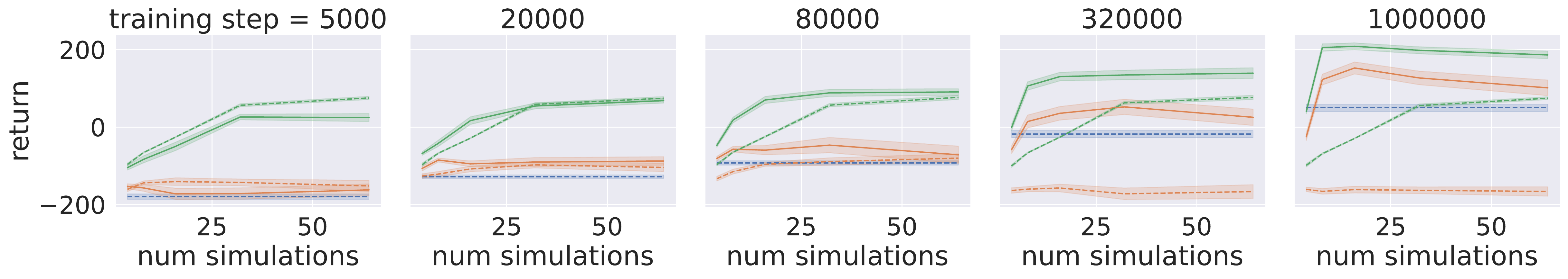}
        \vspace{-15pt}
        \caption{Lunar Lander}
        \vspace{10pt}
    \end{subfigure}
    \begin{subfigure}{0.8\linewidth}
        \centering
        \includegraphics*[width=\textwidth]{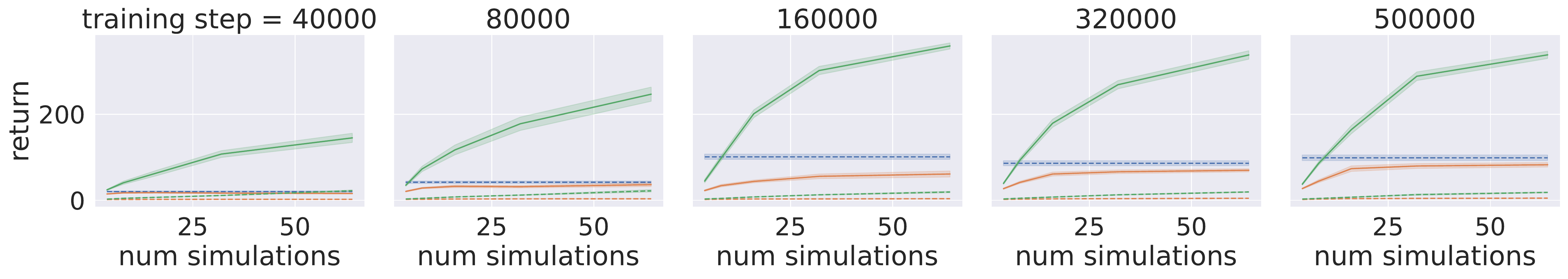}
        \vspace{-15pt}
        \caption{Atari Breakout}
        \vspace{12pt}
    \end{subfigure}
    \vspace{5pt}
    \caption{MCTS planning with \textrm{(i)} MuZero's learned model (orange) and \textrm{(ii)} the ground-truth model (green), under the guidance of \textrm{(a)} MuZero's policy prior (solid line) and \textrm{(b)} the uniform prior (dotted line)). 
    The X and Y axes are the number of simulations per step and the return. }
    \label{fig:policy_improvement}
    \vspace{10pt}
\end{figure*}

We present our results in Figure \ref{fig:policy_improvement}. 
In Cart Pole, MuZero's learned model exhibits some degree of support for free search (orange dashed) after some training. 
However, it significantly underperforms compared to free search with the ground-truth model (green dashed) in terms of both planning efficiency and asymptotic performance.
In Lunar Lander and Atari Breakout, free search with the learned model (orange dashed) fails completely. 
Consequently, the potential for finding a good policy through planning with the learned model will be effectively limited. 

In Cart Pole and Lunar Lander, with enough MCTS simulations, planning with the learned model improves performance over using the policy prior alone. 
However, compared to the ground-truth model, it is evident that the extent to which we can improve upon the policy prior via planning is still very limited. 
This is a clear sign that the model error is restricting the extent to which we can further improve the policy by planning.
In Atari Breakout, while planning with more simulations using the learned model improves the agent's performance, it cannot outperform the policy prior. 

Interestingly, when comparing planning with the policy prior to planning with the uniform prior, the learned model consistently shows a larger gap compared to the ground-truth model except in Atari Breakout. 
Furthermore, unlike the ground-truth model, 
the gap for the learned model does not appear to diminish quickly with more simulations, suggesting an additional role of the policy prior to accelerating MCTS. 
As shown in Figure \ref{fig:value_estimation_error_prob}, action sequences with higher probabilities of being selected by the behavior policy tend to be more accurately evaluated by the learned model. 
As the policy prior is directly learned to match the behavior policy, it is reasonable to assume that the values of action sequences favored by the policy prior can also be more accurately predicted by the learned model. 

\begin{figure}[h]
    \centering
    \includegraphics*[width=\linewidth]{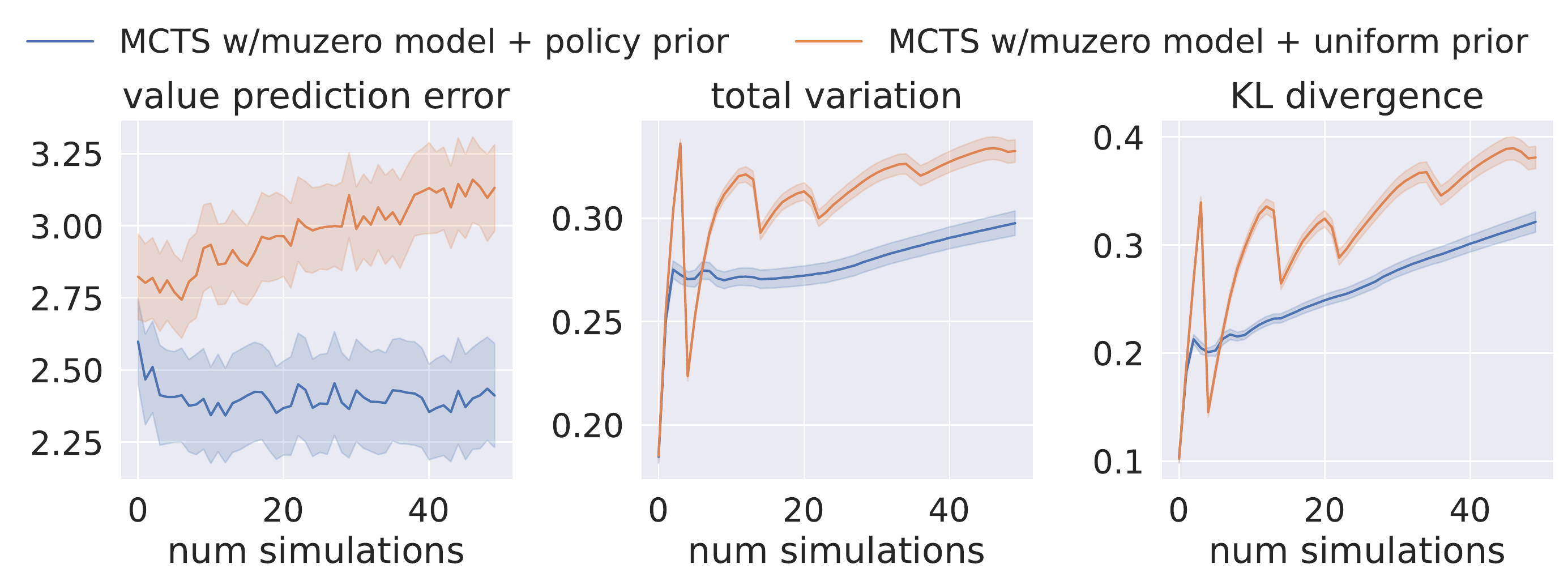}
    \vspace{-5pt}
    \caption{Value prediction error of MCTS's simulated trajectories using the learned model, total variation, and KL divergence (see text).}
    \label{fig:ablation_cartpole}
\end{figure}

To verify this hypothesis, in Figure \ref{fig:ablation_cartpole}, we plot the learned model's value prediction error for MCTS's simulated trajectories under the guidance of the policy prior (blue) and uniform prior (orange) in CartPole after $2000$ training steps.
Meanwhile, we also plot the total variation $\text{TV}[\pi_\theta, \hat{\pi}]$ and KL divergence $\text{KL}[\pi_\theta, \hat{\pi}]$ between the policy prior $\pi_\theta$ and MCTS's empirical visit distribution \(\hat{\pi} = 
\frac{1+N(a)}{\sum_{b}|\mathcal{A}|+N(b)}\) at the root node. 
Clearly, the policy prior regularizes MCTS to visit actions that are more favored by it, as suggested by the lower total variation and KL divergence (blue, middle, and right), which results in a smaller value prediction error (blue, left), when compared to the uniform prior, which explores more out-of-the-box (orange, middle, and right) and incurs more value prediction errors (orange, left).  
This suggests that \emph{apart from biasing the search, the policy prior may also serve to prevent the search from exploring directions where the learned model is less accurate. }
\vspace{-10pt}

\section{Discussion}\label{sec_discussion}
\paragraph{Limitations} In this study, we analyzed MuZero, one of the most successful DMBRL methods that is grounded in the value equivalence principle. 
The conclusions we draw can, therefore, not be directly generalized to other DMBRL methods, certainly not those that employ other auxiliary loss functions on top, e.g., \citep{gelada_deepmdp_2019,van_der_pol_plannable_2020,ye_mastering_2021}. 
However, our analysis clearly shows that, despite its simplicity, using the value equivalence principle does not imply that we learn models that are actually value equivalent. Even when restricting the predictions to its own behavior policy, MuZero's learned model leads to prediction errors that quickly grow with the horizon of prediction.
As such, our study does serve as a warning for other DMBRL methods that aim to use value equivalence as their guiding principle.

\paragraph{Outlook} We have demonstrated that MuZero's model, trained with a value-equivalence-based loss, struggles to generalize and predict values accurately for unseen or unfamiliar policies. 
This raises the question of how different losses for model learning would behave and compare in our analysis, such as the reconstruction-based loss \citep{kaiser_model_2020, hafner_dream_2019,hafner_mastering_2020,hafner_mastering_2023} and the temporal-consistency \citep{van_der_pol_plannable_2020,de_vries_visualizing_2021,ye_mastering_2021}.
Specifically, we hypothesize that, while the value equivalence loss is the most flexible loss as it does not directly impose requirements on state representations, other losses may have an advantage for generalization in low-data regimes due to a richer supervision signal. 
We consider this potential trade-off between representation complexity and generalization an exciting direction for future research. 

\newpage
\section{Conclusion}
In this work, we empirically studied the models learned by MuZero, which are trained based on the value equivalence principle.
Our analysis focused on two fundamental questions: 
\begin{enumerate*}[label=(\arabic*)]
    \item To what extent does MuZero learn a value-equivalent model? and 
    \item To what extent does the learned model support effective policy improvement by planning?
\end{enumerate*} 
We find that MuZero's learned model cannot generally evaluate policies accurately, especially those that further deviate from the data collection policy. 
Consequently, the failure of the model to predict values for policies that are out of the training distribution prevents effective planning from scratch, which limits the extent to which MuZero can additionally improve its policy via planning.
Moreover, we uncover that apart from accelerating the search as in AlphaZero, the policy prior in MuZero serves another crucial function: it regularizes the search towards areas where the learned model is more accurate, which effectively reduces the model error that is accumulated into planning. 
From these findings, we speculate that because MuZero's model itself is limited in its ability for policy improvement, the empirical success of MuZero may be a result of the model providing the algorithm with a more powerful representation of the values and policies compared to single-lookahead methods like deep Q-learning \citep{mnih_human-level_2015}.
This implies a role for the model that is similar to that in model-free planning but extends it with the policy prior to make planning conservative \citep{kumar_conservative_2020}. 

\begin{ack}
The authors acknowledge the use of computational resources provided by the Delft AI Cluster (https://daic.tudelft.nl) and the Delft High Performance Computing Centre (https://www.tudelft.nl/dhpc).
\end{ack}
\clearpage

\bibliography{reference}

\begin{thebibliography}{53}
\providecommand{\natexlab}[1]{#1}
\providecommand{\url}[1]{\texttt{#1}}
\expandafter\ifx\csname urlstyle\endcsname\relax
  \providecommand{\doi}[1]{doi: #1}\else
  \providecommand{\doi}{doi: \begingroup \urlstyle{rm}\Url}\fi

\bibitem[Antonoglou et~al.(2021)Antonoglou, Schrittwieser, Ozair, Hubert, and
  Silver]{antonoglou_planning_2021}
I.~Antonoglou, J.~Schrittwieser, S.~Ozair, T.~K. Hubert, and D.~Silver.
\newblock Planning in {{Stochastic Environments}} with a {{Learned Model}}.
\newblock In \emph{{{ICLR}}}, 2021.

\bibitem[Bellemare et~al.(2013)Bellemare, Veness, and
  Bowling]{bellemare_arcade_2013}
M.~G. Bellemare, J.~Veness, and M.~Bowling.
\newblock The {{Arcade Learning Environment}}: {{An Evaluation Platform}} for
  {{General Agents}}.
\newblock \emph{JAIR}, 2013.

\bibitem[Brafman and Tennenholtz(2002)]{brafman_r-max_2002}
R.~I. Brafman and M.~Tennenholtz.
\newblock R-{{MAX}}-{{A General Polynomial Time Algorithm}} for {{Near-Optimal
  Reinforcement Learning}}.
\newblock \emph{JMLR}, 2002.

\bibitem[Browne et~al.(2012)Browne, Powley, Whitehouse, Lucas, Cowling,
  Rohlfshagen, Tavener, Perez, Samothrakis, and Colton]{browne_survey_2012}
C.~B. Browne, E.~Powley, D.~Whitehouse, S.~M. Lucas, P.~I. Cowling,
  P.~Rohlfshagen, S.~Tavener, D.~Perez, S.~Samothrakis, and S.~Colton.
\newblock A survey of {{Monte Carlo}} tree search methods.
\newblock \emph{IEEE Transactions on Computational Intelligence and AI in
  games}, 2012.

\bibitem[Danihelka et~al.(2021)Danihelka, Guez, Schrittwieser, and
  Silver]{danihelka_policy_2021}
I.~Danihelka, A.~Guez, J.~Schrittwieser, and D.~Silver.
\newblock Policy improvement by planning with {{Gumbel}}.
\newblock In \emph{{{ICLR}}}, 2021.

\bibitem[{de Vries} et~al.(2021){de Vries}, Voskuil, Moerland, and
  Plaat]{de_vries_visualizing_2021}
J.~{de Vries}, K.~Voskuil, T.~M. Moerland, and A.~Plaat.
\newblock Visualizing {{MuZero Models}}.
\newblock In \emph{{{ICML}} 2021 {{Workshop}} on {{Unsupervised Reinforcement
  Learning}}}, 2021.

\bibitem[Degrave et~al.(2022)Degrave, Felici, Buchli, Neunert, Tracey,
  Carpanese, Ewalds, Hafner, Abdolmaleki, {de las Casas}, Donner, Fritz,
  Galperti, Huber, Keeling, Tsimpoukelli, Kay, Merle, and {Moret, et
  al.}]{degrave_magnetic_2022}
J.~Degrave, F.~Felici, J.~Buchli, M.~Neunert, B.~Tracey, F.~Carpanese,
  T.~Ewalds, R.~Hafner, A.~Abdolmaleki, D.~{de las Casas}, C.~Donner, L.~Fritz,
  C.~Galperti, A.~Huber, J.~Keeling, M.~Tsimpoukelli, J.~Kay, A.~Merle, and
  J.-M. {Moret, et al.}
\newblock Magnetic control of tokamak plasmas through deep reinforcement
  learning.
\newblock \emph{Nature}, 2022.

\bibitem[Duvaud and Hainaut(2019)]{muzero-general}
W.~Duvaud and A.~Hainaut.
\newblock {{MuZero}} general: {{Open}} reimplementation of {{MuZero}}.
\newblock https://github.com/werner-duvaud/muzero-general, 2019.

\bibitem[Farquhar et~al.(2018)Farquhar, Rockt{\"a}schel, Igl, and
  Whiteson]{farquhar_treeqn_2018}
G.~Farquhar, T.~Rockt{\"a}schel, M.~Igl, and S.~Whiteson.
\newblock {{TreeQN}} and {{ATreeC}}: {{Differentiable Tree-Structured Models}}
  for {{Deep Reinforcement Learning}}.
\newblock In \emph{{{ICLR}}}, 2018.

\bibitem[{Fran{\c c}ois-Lavet} et~al.(2018){Fran{\c c}ois-Lavet}, Henderson,
  Islam, Bellemare, and Pineau]{francois-lavet_introduction_2018}
V.~{Fran{\c c}ois-Lavet}, P.~Henderson, R.~Islam, M.~G. Bellemare, and
  J.~Pineau.
\newblock An {{Introduction}} to {{Deep Reinforcement Learning}}.
\newblock \emph{Foundations and Trends{\textregistered} in Machine Learning},
  2018.

\bibitem[Gelada et~al.(2019)Gelada, Kumar, Buckman, Nachum, and
  Bellemare]{gelada_deepmdp_2019}
C.~Gelada, S.~Kumar, J.~Buckman, O.~Nachum, and M.~G. Bellemare.
\newblock {{DeepMDP}}: {{Learning Continuous Latent Space Models}} for
  {{Representation Learning}}.
\newblock In \emph{{{ICML}}}, 2019.

\bibitem[Grill et~al.(2020)Grill, Altch{\'e}, Tang, Hubert, Valko, Antonoglou,
  and Munos]{grill_monte-carlo_2020}
J.-B. Grill, F.~Altch{\'e}, Y.~Tang, T.~Hubert, M.~Valko, I.~Antonoglou, and
  R.~Munos.
\newblock Monte-{{Carlo Tree Search}} as {{Regularized Policy Optimization}}.
\newblock In \emph{{{ICML}}}, 2020.

\bibitem[Grimm et~al.(2020)Grimm, Barreto, Singh, and Silver]{grimm_value_2020}
C.~Grimm, A.~Barreto, S.~Singh, and D.~Silver.
\newblock The {{Value Equivalence Principle}} for {{Model-Based Reinforcement
  Learning}}.
\newblock In \emph{{{NeurIPS}}}, 2020.

\bibitem[Grimm et~al.(2021)Grimm, Barreto, Farquhar, Silver, and
  Singh]{grimm_proper_2021}
C.~Grimm, A.~Barreto, G.~Farquhar, D.~Silver, and S.~Singh.
\newblock Proper value equivalence.
\newblock In \emph{{{NeurIPS}}}, 2021.

\bibitem[Grimm et~al.(2022)Grimm, Barreto, and Singh]{grimm_approximate_2022}
C.~Grimm, A.~Barreto, and S.~Singh.
\newblock Approximate {{Value Equivalence}}.
\newblock In \emph{{{NeurIPS}}}, volume~35, 2022.

\bibitem[Guez et~al.(2018)Guez, Weber, Antonoglou, Simonyan, Vinyals, Wierstra,
  Munos, and Silver]{guez_learning_2018}
A.~Guez, T.~Weber, I.~Antonoglou, K.~Simonyan, O.~Vinyals, D.~Wierstra,
  R.~Munos, and D.~Silver.
\newblock Learning to search with {{MCTSnets}}.
\newblock In \emph{{{ICML}}}, 2018.

\bibitem[Guez et~al.(2019)Guez, Mirza, Gregor, Kabra, Racaniere, Weber, Raposo,
  Santoro, Orseau, Eccles, Wayne, Silver, and
  Lillicrap]{guez_investigation_2019}
A.~Guez, M.~Mirza, K.~Gregor, R.~Kabra, S.~Racaniere, T.~Weber, D.~Raposo,
  A.~Santoro, L.~Orseau, T.~Eccles, G.~Wayne, D.~Silver, and T.~Lillicrap.
\newblock An {{Investigation}} of {{Model-Free Planning}}.
\newblock In \emph{{{ICML}}}, 2019.

\bibitem[Ha and Schmidhuber(2018)]{ha_recurrent_2018}
D.~Ha and J.~Schmidhuber.
\newblock Recurrent {{World Models Facilitate Policy Evolution}}.
\newblock In \emph{{{NeurIPS}}}, 2018.

\bibitem[Hafner et~al.(2019)Hafner, Lillicrap, Ba, and
  Norouzi]{hafner_dream_2019}
D.~Hafner, T.~Lillicrap, J.~Ba, and M.~Norouzi.
\newblock Dream to {{Control}}: {{Learning Behaviors}} by {{Latent
  Imagination}}.
\newblock In \emph{{{ICLR}}}, 2019.

\bibitem[Hafner et~al.(2020)Hafner, Lillicrap, Norouzi, and
  Ba]{hafner_mastering_2020}
D.~Hafner, T.~P. Lillicrap, M.~Norouzi, and J.~Ba.
\newblock Mastering {{Atari}} with {{Discrete World Models}}.
\newblock In \emph{{{ICLR}}}, 2020.

\bibitem[Hafner et~al.(2023)Hafner, Pasukonis, Ba, and
  Lillicrap]{hafner_mastering_2023}
D.~Hafner, J.~Pasukonis, J.~Ba, and T.~Lillicrap.
\newblock Mastering {{Diverse Domains}} through {{World Models}}.
\newblock \emph{arXiv preprint arXiv:2301.04104}, 2023.

\bibitem[Hamrick et~al.(2022)Hamrick, Friesen, Behbahani, Guez, Viola,
  Witherspoon, Anthony, Buesing, Veli{\v c}kovi{\'c}, and
  Weber]{hamrick_role_2022}
J.~B. Hamrick, A.~L. Friesen, F.~Behbahani, A.~Guez, F.~Viola, S.~Witherspoon,
  T.~Anthony, L.~H. Buesing, P.~Veli{\v c}kovi{\'c}, and T.~Weber.
\newblock On the role of planning in model-based deep reinforcement learning.
\newblock In \emph{{{ICLR}}}, 2022.

\bibitem[Henaff(2019)]{henaff_explicit_2019}
M.~Henaff.
\newblock Explicit {{Explore-Exploit Algorithms}} in {{Continuous State
  Spaces}}.
\newblock In \emph{{{NeurIPS}}}, 2019.

\bibitem[Hessel et~al.(2022)Hessel, Danihelka, Viola, Guez, Schmitt, Sifre,
  Weber, Silver, and {van Hasselt}]{hessel_muesli_2022}
M.~Hessel, I.~Danihelka, F.~Viola, A.~Guez, S.~Schmitt, L.~Sifre, T.~Weber,
  D.~Silver, and H.~{van Hasselt}.
\newblock Muesli: {{Combining Improvements}} in {{Policy Optimization}}.
\newblock \emph{arXiv preprint arXiv:2104.06159}, 2022.

\bibitem[Hubert et~al.(2021)Hubert, Schrittwieser, Antonoglou, Barekatain,
  Schmitt, and Silver]{hubert_learning_2021}
T.~Hubert, J.~Schrittwieser, I.~Antonoglou, M.~Barekatain, S.~Schmitt, and
  D.~Silver.
\newblock Learning and {{Planning}} in {{Complex Action Spaces}}.
\newblock In \emph{{{ICML}}}, 2021.

\bibitem[Jaderberg et~al.(2022)Jaderberg, Mnih, Czarnecki, Schaul, Leibo,
  Silver, and Kavukcuoglu]{jaderberg_reinforcement_2022}
M.~Jaderberg, V.~Mnih, W.~M. Czarnecki, T.~Schaul, J.~Z. Leibo, D.~Silver, and
  K.~Kavukcuoglu.
\newblock Reinforcement {{Learning}} with {{Unsupervised Auxiliary Tasks}}.
\newblock In \emph{{{ICLR}}}, 2022.

\bibitem[Kaiser et~al.(2020)Kaiser, Babaeizadeh, Mi{\l}os, Osi{\'n}ski,
  Campbell, Czechowski, Erhan, Finn, Kozakowski, Levine, Mohiuddin, Sepassi,
  Tucker, and Michalewski]{kaiser_model_2020}
{\L}.~Kaiser, M.~Babaeizadeh, P.~Mi{\l}os, B.~Osi{\'n}ski, R.~H. Campbell,
  K.~Czechowski, D.~Erhan, C.~Finn, P.~Kozakowski, S.~Levine, A.~Mohiuddin,
  R.~Sepassi, G.~Tucker, and H.~Michalewski.
\newblock Model {{Based Reinforcement Learning}} for {{Atari}}.
\newblock In \emph{{{ICLR}}}, 2020.

\bibitem[Kocsis and Szepesv{\'a}ri(2006)]{kocsis_bandit_2006}
L.~Kocsis and C.~Szepesv{\'a}ri.
\newblock Bandit based {{Monte-Carlo}} planning.
\newblock In \emph{{{ECML}}}, 2006.

\bibitem[Kumar et~al.(2020)Kumar, Zhou, Tucker, and
  Levine]{kumar_conservative_2020}
A.~Kumar, A.~Zhou, G.~Tucker, and S.~Levine.
\newblock Conservative {{Q-Learning}} for {{Offline Reinforcement Learning}}.
\newblock In \emph{{{NeurIPS}}}, 2020.

\bibitem[Lambert et~al.(2022)Lambert, Pister, and
  Calandra]{lambert_investigating_2022}
N.~Lambert, K.~Pister, and R.~Calandra.
\newblock Investigating {{Compounding Prediction Errors}} in {{Learned Dynamics
  Models}}.
\newblock \emph{arXiv preprint arXiv:2203.09637}, 2022.

\bibitem[Li et~al.(2006)Li, Walsh, and Littman]{li_towards_2006}
L.~Li, T.~J. Walsh, and M.~L. Littman.
\newblock Towards a {{Unified Theory}} of {{State Abstraction}} for {{MDPs}}.
\newblock In \emph{ISAIM}, 2006.

\bibitem[Lowrey et~al.(2018)Lowrey, Rajeswaran, Kakade, Todorov, and
  Mordatch]{lowrey_plan_2018}
K.~Lowrey, A.~Rajeswaran, S.~Kakade, E.~Todorov, and I.~Mordatch.
\newblock Plan {{Online}}, {{Learn Offline}}: {{Efficient Learning}} and
  {{Exploration}} via {{Model-Based Control}}.
\newblock In \emph{{{ICLR}}}, 2018.

\bibitem[Madeka et~al.(2022)Madeka, Torkkola, Eisenach, Luo, Foster, and
  Kakade]{madeka_deep_2022}
D.~Madeka, K.~Torkkola, C.~Eisenach, A.~Luo, D.~P. Foster, and S.~M. Kakade.
\newblock Deep {{Inventory Management}}.
\newblock \emph{arXiv preprint arXiv:2210.03137}, 2022.

\bibitem[Mandhane et~al.(2022)Mandhane, Zhernov, Rauh, Gu, Wang, Xue, Shang,
  Pang, Claus, Chiang, Chen, Han, Chen, Mankowitz, Broshear, Schrittwieser,
  Hubert, Vinyals, and Mann]{mandhane_muzero_2022}
A.~Mandhane, A.~Zhernov, M.~Rauh, C.~Gu, M.~Wang, F.~Xue, W.~Shang, D.~Pang,
  R.~Claus, C.-H. Chiang, C.~Chen, J.~Han, A.~Chen, D.~J. Mankowitz,
  J.~Broshear, J.~Schrittwieser, T.~Hubert, O.~Vinyals, and T.~Mann.
\newblock {{MuZero}} with {{Self-competition}} for {{Rate Control}} in {{VP9
  Video Compression}}.
\newblock \emph{arXiv preprint arXiv:2202.06626}, 2022.

\bibitem[McCallum(1996)]{mccallum_reinforcement_1996}
A.~K. McCallum.
\newblock \emph{Reinforcement Learning with Selective Perception and Hidden
  State}.
\newblock PhD thesis, University of Rochester, 1996.

\bibitem[Mirhoseini et~al.(2021)Mirhoseini, Goldie, Yazgan, Jiang, Songhori,
  Wang, Lee, Johnson, Pathak, Nazi, Pak, Tong, Srinivasa, Hang, Tuncer, Le,
  Laudon, Ho, and {Carpenter, et al.}]{mirhoseini_graph_2021}
A.~Mirhoseini, A.~Goldie, M.~Yazgan, J.~W. Jiang, E.~Songhori, S.~Wang, Y.-J.
  Lee, E.~Johnson, O.~Pathak, A.~Nazi, J.~Pak, A.~Tong, K.~Srinivasa, W.~Hang,
  E.~Tuncer, Q.~V. Le, J.~Laudon, R.~Ho, and R.~{Carpenter, et al.}
\newblock A graph placement methodology for fast chip design.
\newblock \emph{Nature}, 2021.

\bibitem[Mnih et~al.(2015)Mnih, Kavukcuoglu, Silver, Rusu, Veness, Bellemare,
  Graves, Riedmiller, Fidjeland, Ostrovski, Petersen, Beattie, Sadik,
  Antonoglou, King, Kumaran, Wierstra, Legg, and
  Hassabis]{mnih_human-level_2015}
V.~Mnih, K.~Kavukcuoglu, D.~Silver, A.~A. Rusu, J.~Veness, M.~G. Bellemare,
  A.~Graves, M.~Riedmiller, A.~K. Fidjeland, G.~Ostrovski, S.~Petersen,
  C.~Beattie, A.~Sadik, I.~Antonoglou, H.~King, D.~Kumaran, D.~Wierstra,
  S.~Legg, and D.~Hassabis.
\newblock Human-level control through deep reinforcement learning.
\newblock \emph{Nature}, 2015.

\bibitem[Moerland et~al.(2023)Moerland, Broekens, Plaat, and
  Jonker]{moerland_model-based_2023}
T.~M. Moerland, J.~Broekens, A.~Plaat, and C.~M. Jonker.
\newblock Model-based {{Reinforcement Learning}}: {{A Survey}}.
\newblock \emph{Foundations and Trends{\textregistered} in Machine Learning},
  2023.

\bibitem[Oh et~al.(2017)Oh, Singh, and Lee]{oh_value_2017}
J.~Oh, S.~Singh, and H.~Lee.
\newblock Value {{Prediction Network}}.
\newblock In \emph{{{NeurIPS}}}, 2017.

\bibitem[Pathak et~al.(2017)Pathak, Agrawal, Efros, and
  Darrell]{pathak_curiosity-driven_2017}
D.~Pathak, P.~Agrawal, A.~A. Efros, and T.~Darrell.
\newblock Curiosity-driven {{Exploration}} by {{Self-supervised Prediction}}.
\newblock \emph{arXiv preprint arXiv:1705.05363}, 2017.

\bibitem[Pathak et~al.(2019)Pathak, Gandhi, and
  Gupta]{pathak_self-supervised_2019}
D.~Pathak, D.~Gandhi, and A.~Gupta.
\newblock Self-{{Supervised Exploration}} via {{Disagreement}}.
\newblock In \emph{{{ICML}}}, 2019.

\bibitem[Schrittwieser et~al.(2020)Schrittwieser, Antonoglou, Hubert, Simonyan,
  Sifre, Schmitt, Guez, Lockhart, Hassabis, and
  Graepel]{schrittwieser_mastering_2020}
J.~Schrittwieser, I.~Antonoglou, T.~Hubert, K.~Simonyan, L.~Sifre, S.~Schmitt,
  A.~Guez, E.~Lockhart, D.~Hassabis, and T.~Graepel.
\newblock Mastering atari, go, chess and shogi by planning with a learned
  model.
\newblock \emph{Nature}, 2020.

\bibitem[Sekar et~al.(2020)Sekar, Rybkin, Daniilidis, Abbeel, Hafner, and
  Pathak]{sekar_planning_2020}
R.~Sekar, O.~Rybkin, K.~Daniilidis, P.~Abbeel, D.~Hafner, and D.~Pathak.
\newblock Planning to {{Explore}} via {{Self-Supervised World Models}}.
\newblock In \emph{{{ICML}}}, 2020.

\bibitem[Shyam et~al.(2019)Shyam, Ja{\'s}kowski, and
  Gomez]{shyam_model-based_2019}
P.~Shyam, W.~Ja{\'s}kowski, and F.~Gomez.
\newblock Model-{{Based Active Exploration}}.
\newblock In \emph{{{ICML}}}, 2019.

\bibitem[Silver et~al.(2016)Silver, Huang, Maddison, Guez, Sifre, Van
  Den~Driessche, Schrittwieser, Antonoglou, Panneershelvam, Lanctot, Dieleman,
  Grewe, Nham, Kalchbrenner, Sutskever, Lillicrap, Leach, Kavukcuoglu, and
  {Graepel, et al.}]{silver_mastering_2016}
D.~Silver, A.~Huang, C.~J. Maddison, A.~Guez, L.~Sifre, G.~Van Den~Driessche,
  J.~Schrittwieser, I.~Antonoglou, V.~Panneershelvam, M.~Lanctot, S.~Dieleman,
  D.~Grewe, J.~Nham, N.~Kalchbrenner, I.~Sutskever, T.~Lillicrap, M.~Leach,
  K.~Kavukcuoglu, and T.~{Graepel, et al.}
\newblock Mastering the game of {{Go}} with deep neural networks and tree
  search.
\newblock \emph{Nature}, 2016.

\bibitem[Silver et~al.(2017{\natexlab{a}})Silver, Hasselt, Hessel, Schaul,
  Guez, Harley, {Dulac-Arnold}, Reichert, Rabinowitz, Barreto, and
  Degris]{silver_predictron_2017}
D.~Silver, H.~Hasselt, M.~Hessel, T.~Schaul, A.~Guez, T.~Harley,
  G.~{Dulac-Arnold}, D.~Reichert, N.~Rabinowitz, A.~Barreto, and T.~Degris.
\newblock The {{Predictron}}: {{End-To-End Learning}} and {{Planning}}.
\newblock In \emph{{{ICML}}}, 2017{\natexlab{a}}.

\bibitem[Silver et~al.(2017{\natexlab{b}})Silver, Schrittwieser, Simonyan,
  Antonoglou, Huang, Guez, Hubert, Baker, Lai, Bolton, Chen, Lillicrap, Hui,
  Sifre, Van Den~Driessche, Graepel, and Hassabis]{silver_mastering_2017}
D.~Silver, J.~Schrittwieser, K.~Simonyan, I.~Antonoglou, A.~Huang, A.~Guez,
  T.~Hubert, L.~Baker, M.~Lai, A.~Bolton, Y.~Chen, T.~Lillicrap, F.~Hui,
  L.~Sifre, G.~Van Den~Driessche, T.~Graepel, and D.~Hassabis.
\newblock Mastering the game of {{Go}} without human knowledge.
\newblock \emph{Nature}, 2017{\natexlab{b}}.

\bibitem[Silver et~al.(2018)Silver, Hubert, Schrittwieser, Antonoglou, Lai,
  Guez, Lanctot, Sifre, Kumaran, Graepel, Lillicrap, Simonyan, and
  Hassabis]{silver_general_2018}
D.~Silver, T.~Hubert, J.~Schrittwieser, I.~Antonoglou, M.~Lai, A.~Guez,
  M.~Lanctot, L.~Sifre, D.~Kumaran, T.~Graepel, T.~Lillicrap, K.~Simonyan, and
  D.~Hassabis.
\newblock A general reinforcement learning algorithm that masters chess, shogi,
  and {{Go}} through self-play.
\newblock \emph{Science}, 2018.

\bibitem[Sutton(1991)]{sutton_dyna_1991}
R.~S. Sutton.
\newblock Dyna, an integrated architecture for learning, planning, and
  reacting.
\newblock \emph{ACM SIGART Bulletin}, 1991.

\bibitem[Tamar et~al.(2016)Tamar, WU, Thomas, Levine, and
  Abbeel]{tamar_value_2016}
A.~Tamar, {\relax YI}.~WU, G.~Thomas, S.~Levine, and P.~Abbeel.
\newblock Value {{Iteration Networks}}.
\newblock In \emph{{{NeurIPS}}}, 2016.

\bibitem[{van der Pol} et~al.(2020){van der Pol}, Kipf, Oliehoek, and
  Welling]{van_der_pol_plannable_2020}
E.~{van der Pol}, T.~Kipf, F.~A. Oliehoek, and M.~Welling.
\newblock Plannable {{Approximations}} to {{MDP Homomorphisms}}:
  {{Equivariance}} under {{Actions}}.
\newblock In \emph{{{AAMAS}}}, 2020.

\bibitem[{van Hasselt} et~al.(2019){van Hasselt}, Hessel, and
  Aslanides]{van_hasselt_when_2019}
H.~P. {van Hasselt}, M.~Hessel, and J.~Aslanides.
\newblock When to use parametric models in reinforcement learning?
\newblock In \emph{{{NeurIPS}}}, 2019.

\bibitem[Ye et~al.(2021)Ye, Liu, Kurutach, Abbeel, and Gao]{ye_mastering_2021}
W.~Ye, S.~Liu, T.~Kurutach, P.~Abbeel, and Y.~Gao.
\newblock Mastering {{Atari Games}} with {{Limited Data}}.
\newblock In \emph{{{NeurIPS}}}, 2021.

\end{thebibliography}

\newpage
\onecolumn

\appendix
\section{Training of MuZero agents}\label{appendix:training_agents}

\subsection{Setup and Hyperparameters}

\begin{table}[h]
  \centering
  \begin{tabular}{|l|l|l|}
    \hline
    \textbf{Hyperparameter} & \textbf{Cart Pole} & \textbf{Lunar Lander} \\ \hline
    Random seeds & 0 to 29 & 0 to 29 \\
    Discount factor & 0.997 & 0.999 \\
    Total training steps & \textbf{100000} (10000) & \textbf{1000000} (200000) \\
    Optimizer & Adam & Adam \\
    Initial Learning Rate & 0.02 & 0.005 \\
    Learning Rate Decay Rate & \textbf{0.1} (0.8) & No decay  \\
    Learning Rate Decay Steps & \textbf{50000} (1000) & No decay \\
    Weight Decay & 1e-4 & 1e-4 \\
    Momentum & 0.9 & 0.9 \\
    Batch Size & 128 & 64 \\
    Encoding size & 8 & 10 \\
    Fully-connected Layer Size & 16 & 64 \\
    Root Dirichlet Alpha & 0.25 & 0.25 \\
    Root Dirichlet Fraction & 0.25 & 0.25 \\
    Prioritized Experience Replay Alpha & 0.5 & 0.5 \\
    Num Unroll Steps & 10 & 10 \\
    TD Steps & 50 & 30 \\
    Support Size & 10 & 10 \\
    Value Loss Weight & 1.0 & 1.0 \\
    Replay Buffer Size & 500 & \textbf{500} (2000) \\
    Visit Softmax Temperature Fn & 1.0 $\rightarrow$ (5e4) 0.5 $\rightarrow$ (7.5e4)  0.25 & 0.35 \\
    \hline
  \end{tabular}
  \vspace{5pt}
  \caption{Hyperparameters for training MuZero agents in Cart Pole and Lunar Lander. We used default values from \citep{muzero-general} for most of the hyperparameters. The bold values are those that we tuned to improve the convergence of the agents, with the default values shown in brackets.}
  \label{tab:hyperparameters}
\end{table}

\begin{table}[h]
  \centering
  \begin{tabular}{|l|l|l|}
    \hline
    \textbf{Hyperparameter} & \textbf{Atari Breakout} \\ \hline
    Random seeds & 0 to 9 \\
    Total training steps & 500000 \\
    Learning Rate & $0.1 \rightarrow 0.01$ with an exponential decay rate of $0.1$ \\
    Replay Buffer Size & 100000  \\
    Visit Softmax Temperature Fn & 1.0 (fixed throughout training) \\
    \hline
  \end{tabular}
  \vspace{5pt}
  \caption{Hyperparameters for training MuZero agents in Atari Breakout. We used default values from EfficientZero \citep{ye_mastering_2021} (see Appendix A.1 Table 6 of \citep{ye_mastering_2021}) for those hyperparameters that are not mentioned in the table. }
  \label{tab:hyperparameters_atari}
\end{table}

For Cart Pole and Lunar Lander, we trained 30 MuZero agents using an open-source implementation of MuZero \citep{muzero-general}, which is available on GitHub (under the MIT license). The implementation has been extensively tested on various classic RL environments, including Cart Pole and Lunar Lander. While we mostly used the default recommended hyperparameter values from \citep{muzero-general}, we fine-tuned a few of them to improve the convergence of agents. The comprehensive list of hyperparameters for training MuZero agents in these environments can be found in Table \ref{tab:hyperparameters}. For Atari Breakout, we trained 20 MuZero agents using the official implementation of EfficientZero \citep{ye_mastering_2021}, \emph{excluding} the additional improvements introduced by EfficientZero. This implementation is also available on GitHub (under the GPL-3.0 license). At the time of this research, this was the only plausible way to train MuZero agents in Atari games.
See Table \ref{tab:hyperparameters_atari} for the hyperparameters.

Regarding computation, training each MuZero agent took around 3 hours for Cart Pole and 20 hours for Lunar Lander, using 8 CPUs. For Atari Breakout, it took around 40 hours per agent using 2 GPUs and 48 CPUs. The training was conducted on a shared internal cluster equipped with a variety of CPUs and GPUs (Nvidia 2080Ti/V100/A40).

\subsection{Learning Curves}
\begin{figure}[h]
    \centering
    \includegraphics[width=0.65\linewidth]{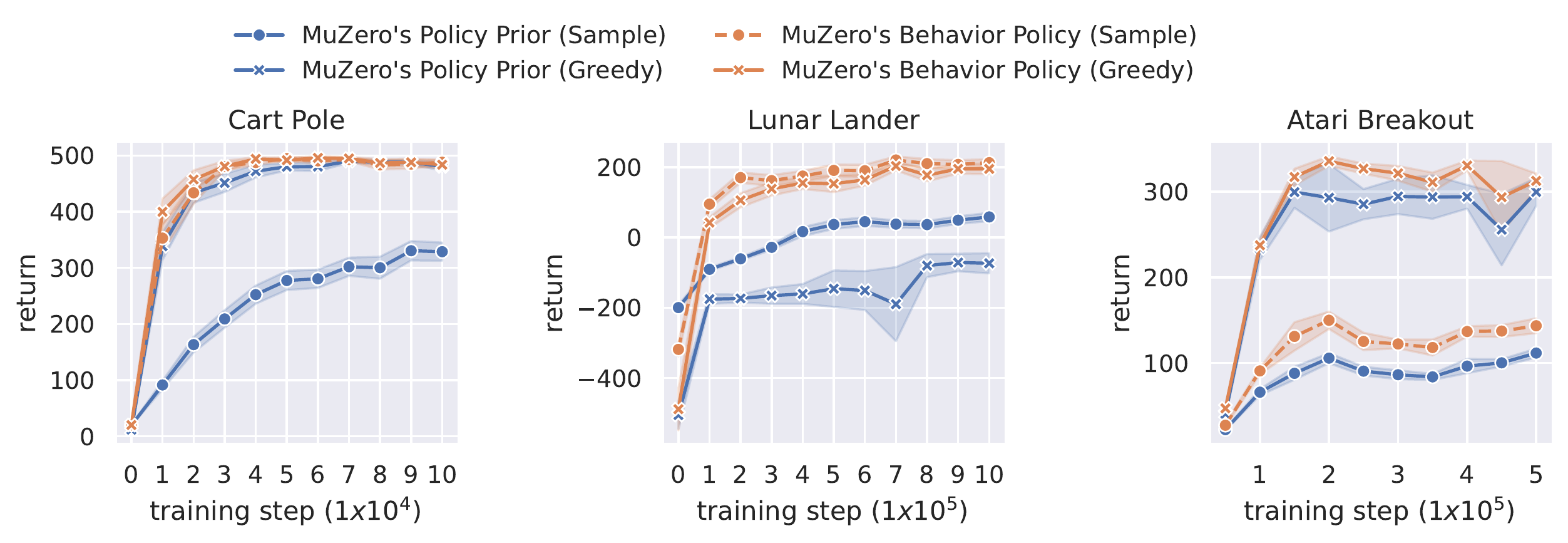}
    \vspace{0pt}
    \caption{Full learning curves of MuZero agents in Cart Pole (Left), Lunar Lander (Middle), and Atari Breakout (Right).}
    \vspace{15pt}
    \label{fig:learning_all}
\end{figure}

In Figure \ref{fig:training}, we plot the online performance of MuZero agents at various training steps for running both the policy prior and the behavior policy. 
There, actions are sampled from policies.
In Figure \ref{fig:learning_all}, we plot the full learning curves with actions both sampled and taken greedily from the policies. When running MuZero's behavior policy greedily, we do not add Dirichlet noise to the policy prior in the tree search. 

\section{Additional Results}

\subsection{Policy Evaluation Experiments with Longer Evaluation Horizons}

In the main paper, we used relatively short evaluation horizons in the policy evaluation experiments because the models were only unrolled for these numbers of steps during training: ${10,10,5}$ for Cart Pole, Lunar Lander, and Atari Breakout, respectively. 
Here, we present results using a longer evaluation horizon of $50$ steps.

\begin{figure*}[h]
    \centering
    \begin{subfigure}[t]{0.33\linewidth}
        \centering
        \includegraphics*[width=\linewidth]{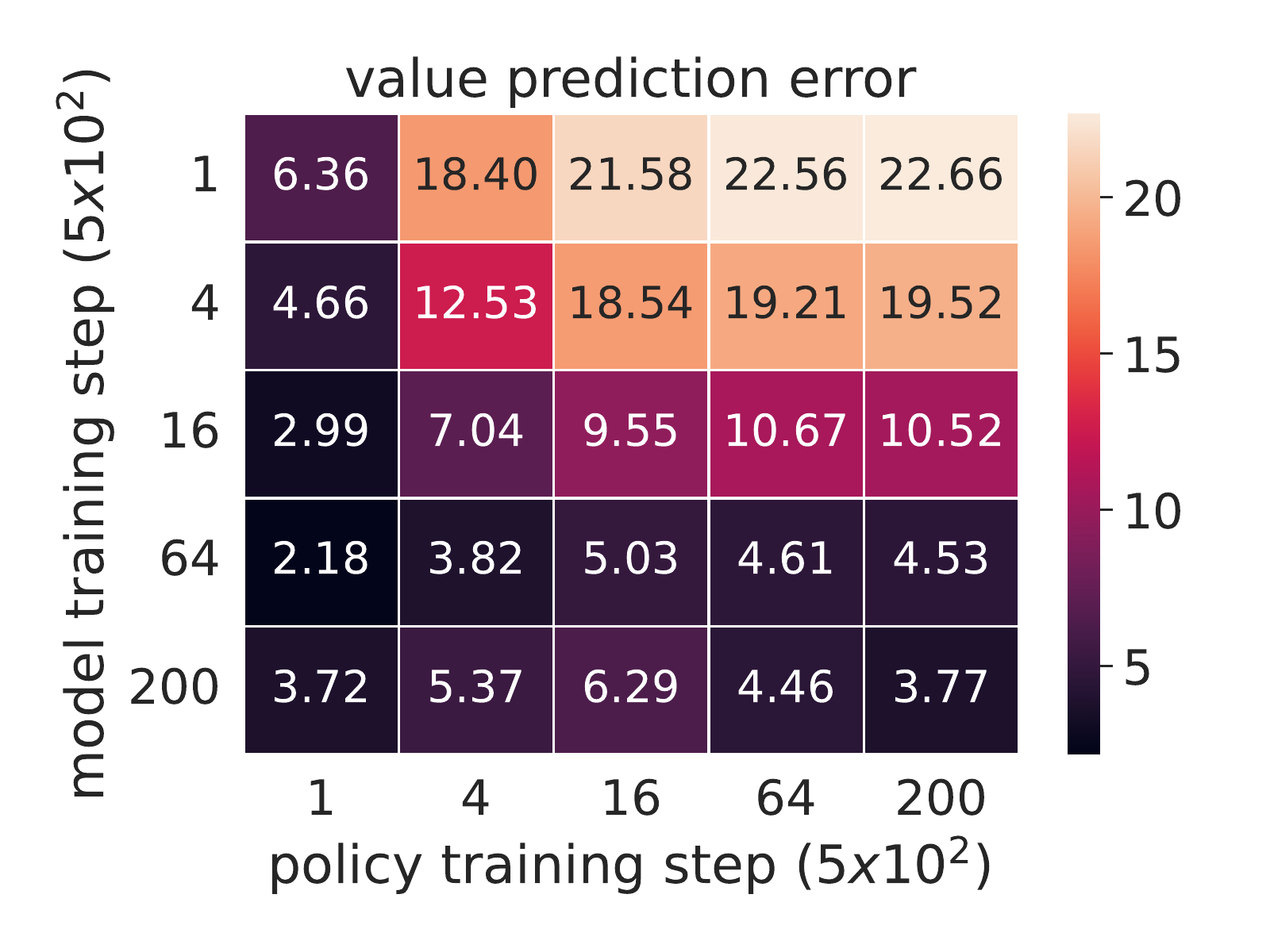}
         \vspace{-15pt}
        \caption{Cart Pole}
    \end{subfigure}
    \begin{subfigure}[t]{0.33\linewidth}
        \centering
        \includegraphics*[width=\linewidth]{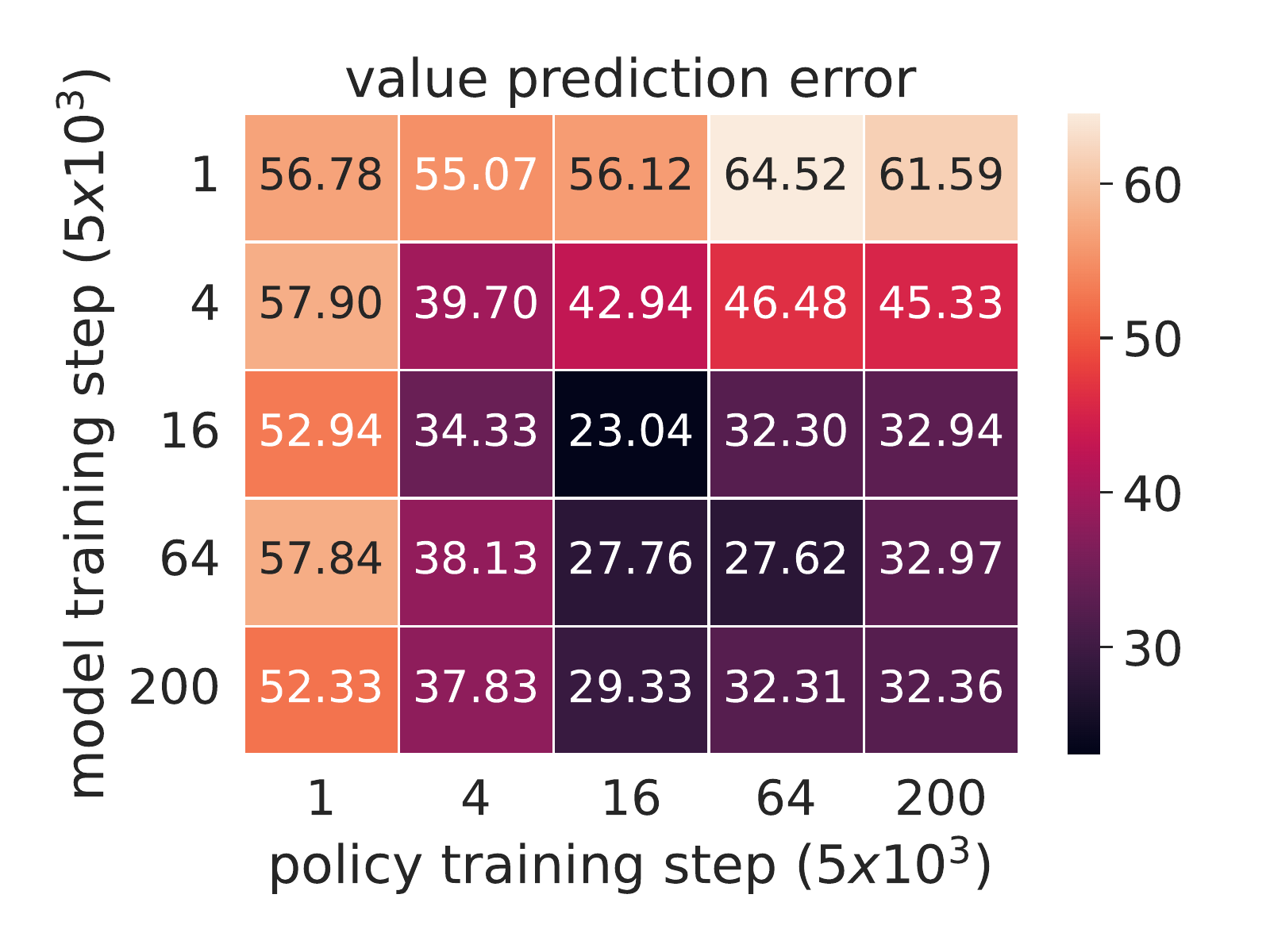}
        \vspace{-15pt}
        \caption{Lunar Lander}
    \end{subfigure}
    \begin{subfigure}[t]{0.33\linewidth}
        \centering
        \includegraphics*[width=\linewidth]{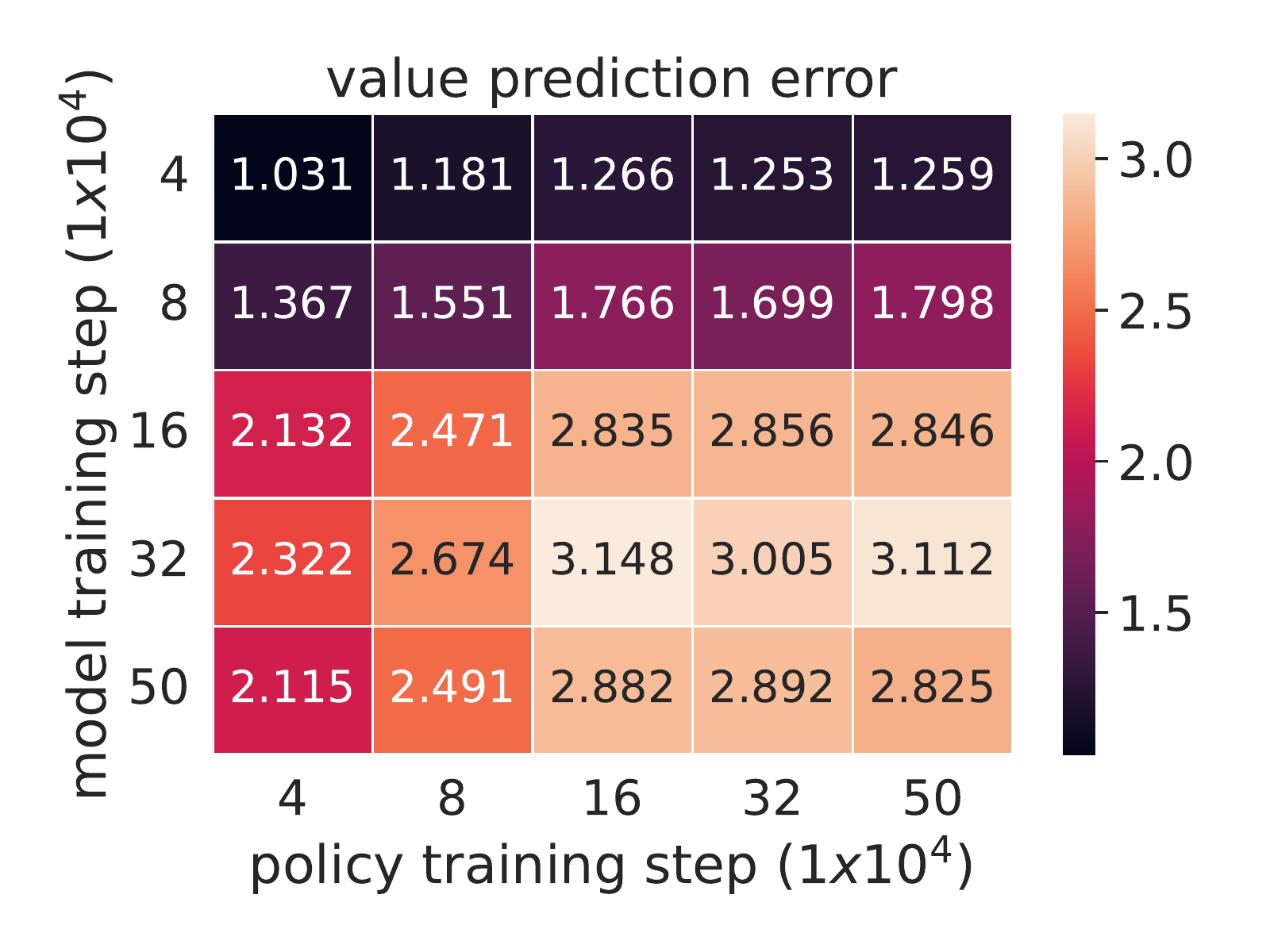}
         \vspace{-15pt}
    \caption{Atari Breakout}
    \vspace{15pt}
    \end{subfigure}
        \caption{Cross model policy evaluation with an evaluation horizon of $50$. We evaluate MuZero's behavior policy at training step Y (column) with the learned model at training step X (row) and measure the value prediction error. Results are aggregated over states sampled from MuZero's on-policy state distribution at training step X (same as the model). 
    }
    \vspace{10pt}
    \label{fig:cross_model_policy_evaluation_longer_horizon}
\end{figure*}

Specifically, we replicate the experiment from Section \ref{sec:cross_model_policy_eval} and Figure \ref{fig:cross_model_policy_evaluation}, where we measure the value prediction error using the model learned at training step $X$ to evaluate MuZero's behavior policy at training step $Y$, but with an extended evaluation horizon of $50$ steps across all domains. The results are shown in Figure \ref{fig:cross_model_policy_evaluation_longer_horizon}.

Compared to Figure \ref{fig:cross_model_policy_evaluation}, which uses shorter evaluation horizons, the value prediction errors here are significantly larger due to compounding errors. However, the overall trend remains consistent: models and policies at different training steps generally show incompatibility in value prediction. This trend is more pronounced here, as the value discrepancies between different policies are more evident with a longer horizon.

\end{document}